\documentclass[preprint,12pt]{elsarticle}

\usepackage{graphicx}
\usepackage{multirow}
\usepackage{array}
\usepackage{tabularx}
\usepackage{booktabs}
\usepackage{amsmath,amssymb}
\usepackage[dvipsnames]{xcolor}
\usepackage{enumitem}
\usepackage{microtype}
\usepackage{url}

\usepackage{pgfplots}
\usepgfplotslibrary{groupplots}
\usetikzlibrary{shapes.geometric,arrows.meta,positioning,calc}
\definecolor{execfill}{HTML}{EEF2F7}
\definecolor{decidefill}{HTML}{FDE9D0}
\definecolor{genfill}{HTML}{E6F2E6}
\newcolumntype{Y}[1]{>{\raggedright\arraybackslash\hsize=#1\hsize}X}
\pgfplotsset{compat=1.18}

\newcolumntype{L}{>{\raggedright\arraybackslash}X}
\setlist[itemize]{leftmargin=1.4em,itemsep=1pt,topsep=2pt}
\setlist[enumerate]{leftmargin=1.6em,itemsep=1pt,topsep=2pt}

\newcommand{\code}[1]{\texttt{\small #1}}

\journal{Knowledge-Based Systems}

\begin{document}

\begin{frontmatter}

\title{Compiling Agent Traces into Typed Workflows for Cheaper and Faster Execution of Recurring
Tasks}

\author[ua]{Jorge Garc\'ia-Carrasco\corref{cor1}}
\ead{jorge.g@ua.es}
\author[ua]{Javier Sanchis}
\ead{javier.sanchis@ua.es}
\author[ua]{Sandra Garc\'ia-Ponsoda}
\ead{sandra.gp@ua.es}
\author[ua]{Alejandro Mat\'e}
\ead{amate@dlsi.ua.es}
\author[ua]{Juan Trujillo}
\ead{jtrujillo@dlsi.ua.es}

\affiliation[ua]{organization={Department of Software and Computing Systems, University of
Alicante}, city={Alicante}, postcode={03690}, country={Spain}}

\cortext[cor1]{Corresponding author.}

\begin{abstract}
LLM agents solve a wide range of digital tasks, and benchmarks now report high success on requests
that span several applications and many steps. However, much of that work is repetitive: a user asks
for the same kind of change again and again, such as moving an alarm or rescheduling a meeting, with
different data but a familiar procedure. Running the full agent for every instance is therefore
expensive.
To this aim, we propose an intermediate representation for such workflows: a directed graph whose
nodes fall into three categories by output space, EXECUTE for computation and tool calls, DECIDE for
one bounded choice over candidates the graph declares, and GENERATE for the open-ended rest. The idea
is that a compiler agent takes the successful traces and compiles them into such a representation.
An engine then runs the graph, calling a model only where a node declares one, so a deployment keeps
neither the agent loop nor its context.
We evaluate on AppWorld, compiling from one agent's traces and running the result on held-out
instances. We show that the compiled workflows solved all instances at 98 times lower
cost and 76 times faster per run.
Compiling a workflow took three to five minutes and paid for itself after 31 to 50
executions.
\end{abstract}


\end{frontmatter}

\section{Introduction}
\label{sec:intro}
Large Language Models (LLMs) are neural networks based on the Transformer architecture
\citep{vaswani2017attention} that are trained on broad amounts of data in a self-supervised manner
\citep{brown2020language}. These models have shown impressive performance across a wide range of
tasks, and after instruction tuning they are able to follow natural language requests directly
\citep{ouyang2022training}. More recently, LLMs have been used as the reasoning core of agents that
interleave reasoning with calls to external tools and APIs \citep{yao2023react, schick2023toolformer},
which allows them to perform actions in external systems rather than only generate text. This has
made LLM agents a practical approach for everyday digital tasks such as managing a calendar,
ordering products or updating the records of a system, and benchmarks now measure this ability across apps \citep{trivedi2024appworld},
web environments \citep{zhou2024webarena}, software repositories \citep{jimenez2024swebench} and
multiple environments at once \citep{liu2024agentbench}.

However, many of the tasks that these agents perform are not new. Users often ask for the same kind
of operation repeatedly, over instances that differ in a few values but share the same underlying
workflow. For example, a personal assistant that manages the alarms of a phone may be asked every
week to move the wake-up alarm half an hour later and disable the rest, because the following day
is a holiday. The specific alarm, the direction and amount of the shift and the day change from one
request to the next, whereas the procedure remains the same: read the alarms, identify the one the
user refers to, compute the new time, set it, disable the others and confirm the change.

A basic LLM agent solves each of these requests from scratch, since nothing from a previous execution is
available to it except the current conversation. Therefore, at every request the agent explores the
available operations, reads their documentation, decides which one to call, checks the result and
corrects itself when the result is not the expected one. This flexibility allows the agent to absorb
the variation between instances, but it also means that the whole procedure is paid for at every
request. On the other hand, a deterministic script would capture the underlying procedure at a negligible cost, but it
is likely to break on the first paraphrase that names a different alarm; an agent absorbs such
paraphrase, but pays for the entire procedure again.

A successful execution leaves a trace, which records the agent's reasoning, the
operations it called with their arguments and the results it obtained. Existing work has used
earlier executions to reduce the cost of recurring tasks, by storing information across interactions
in memory systems \citep{packer2023memgpt}, by inducing reusable skills from successful traces
\citep{wang2023voyager, wang2024awm, skilldisco2026}, or, most recently, by compiling traces into
workflows, programs and state machines \citep{tracecompiler2026, flashscript2026, preact2026}.
These systems show that compilation is possible. What they leave open is the form of the artifact
and what it costs: in each of them a model call returns a value like any other, what the model may
answer is not a closed set of candidates that the artifact declares, and none reports what it cost
to compile the artifact and run it against the agent it replaces. We ask what such an artifact
should contain: which steps of a task can be lowered into code, which need a bounded decision, and
which need open-ended generation, and what the result costs to compile and to run.

In this work, we hypothesize that such a representation can be built from three types of nodes,
defined by the space of values that each step produces. EXECUTE nodes perform deterministic
operations, such as calling an API or computing a value from data at hand, and require no model.
DECIDE nodes perform a bounded semantic decision, that is, a choice among a set of candidates that
the system can enumerate, and can be executed by a typed decision model \citep{jev2026}.
GENERATE
nodes produce free-form values whose space cannot be enumerated and require a generative model,
acting as an escape hatch for the parts of a task that cannot be reduced further. We compile the
traces into a directed graph of such nodes and study whether this representation is sufficient to
execute the task, and which steps end up in each category.

To this end, we propose a compiler that receives the traces of a task, a contract that describes how
the resulting artifact is executed, and a live environment that contains only the traced instances.
The compiler writes the graph and the operations that it references, and iterates by executing the
artifact that it has written. The task itself is never described to the compiler, so any knowledge
about the task that the workflow contains comes from the traces.

We evaluate the approach on AppWorld \citep{trivedi2024appworld}, a benchmark of everyday tasks over
nine apps, with three task families that differ in how much of the request the system can resolve by
itself. For each family, we compile the workflows from a set of successful traces and evaluate them
on held-out instances of the same requests that were never traced. The results show that the
compiled workflows solve the held-out instances at a small fraction of the cost and latency of the
agent, and the analysis of the compiled graphs explains where this reduction comes from: in one
family the procedure is fully resolved into deterministic operations, whereas the others require a
small number of bounded decisions per execution. We also study how many traces are needed, and
whether independent compilations of the same traces produce the same workflow.

Work that couples language models with knowledge-based methods has shown that a model can instantiate
structured knowledge from text \citep{ciatto2025oracles} and that the combination helps where a task
needs both \citep{yang2025survey}. The artifact studied here is such a structure: it states which
operations the task needs, which choices the system can bound and which parts remain open, and each
node declares who may execute it and how its result is checked.

The artifact also has a property that the agent does not have: it can be inspected before it runs. The graph makes explicit which operations are performed, which steps require a
semantic decision and which branch follows from each answer, and the artifact is validated against
its contract before execution. This does not make the workflow correct, but it makes the parts that
can be reviewed, tested or restricted explicit in the artifact instead of implicit in the prompt.

In summary, the contributions of this work are:
\begin{itemize}
  \item \textbf{An intermediate representation of compiled workflows.} We define EXECUTE, DECIDE and
  GENERATE over the space of values that a step produces, derive what may execute and check a node
  in each category, and show that the partition is exhaustive with respect to what the artifact can
  express (Section~\ref{sec:ir}).
  \item \textbf{A compiler from traces to executable workflows.} The compiler receives the traces,
  the IR specification, an empty directory and a world restricted to the traced instances, and writes a
  package that encodes the compiled workflow and can be executed (Section~\ref{sec:compilation}).
  \item \textbf{An empirical study of compiled workflows on a benchmark of everyday tasks.} We
  compile three task families of AppWorld and compare the resulting workflows with the agent that
  produced the traces on held-out instances, in success, cost, latency and model calls (Section~\ref{sec:results}).
  \item \textbf{An analysis of what can be compiled and at what cost.} We study which steps of each
  task are resolved into deterministic operations and which remain semantic decisions, how many
  traces are required, and whether independent compilations agree; from these results we derive
  when compiling a task pays off (Sections~\ref{sec:results} and~\ref{sec:discussion}).
\end{itemize}

The remainder of the paper is structured as follows. Section~\ref{sec:related} reviews recurring
process automation and programming by demonstration, programs generated as an execution substrate,
skill induction from agent traces, the benchmarks we use and cost-aware inference.
Section~\ref{sec:ir} presents the IR and its consequences, and Section~\ref{sec:compilation} the
compiler and the ambiguities it exposed. Section~\ref{sec:setup} describes the benchmark, the split,
the agents and the cost accounting. Section~\ref{sec:results} reports the comparison, the three graphs, the coverage curve, the
stability check and the ablations; Section~\ref{sec:discussion} discusses when compilation pays,
what lowers and where the saving comes from.
Finally, the conclusions are presented in Section~\ref{sec:conclusion}.
 \section{Related Work}
\label{sec:related}
\subsection{From recorded work to compiled artifacts}
Two older lines produce a program where an agent would call a model at every step. Robotic process
automation records a sequence of interface actions once and replays it \citep{vanderAalst2018rpa};
process mining recovers a process model from logs of completed executions, the closest classical
analogue of reading a procedure from a trace \citep{vitale2025controlflow}; programming by
demonstration learns the procedure from examples instead, as FlashFill does for
string transformations \citep{gulwani2011flashfill} and DreamCoder for reusable abstractions
\citep{ellis2021dreamcoder}. A third line writes the program at inference time: PAL and Program of
Thoughts have a model emit code and leave the computation to an interpreter
\citep{gao2023pal, chen2023pot}, ViperGPT composes vision modules with generated Python
\citep{suris2023vipergpt}, and Code as Policies generates robot policy code
\citep{liang2023codeaspolicies}. HuggingGPT has a model plan a task and dispatch each subtask to a
model it selects \citep{shen2023hugginggpt}. All three share the premise we start from, that some steps of a
task are better executed than inferred. Their input, however, is a recording of fixed actions, a
set of value pairs or the instance in hand, and none of them takes a trace of tool calls as its
input or has to separate the procedure of a task from the observations of one instance. When the
program is written at inference time, moreover, a model call stays on the path of every execution.

A fourth line generates the workflow itself, without a recording or an instance at hand. AFlow
searches over code-represented agent workflows and refines them with execution feedback
\citep{zhang2025aflow}; WorkflowLLM fine-tunes a model to write workflows from instructions
\citep{fan2024workflowllm}; LLMCompiler compiles a plan into a graph of function calls that it
dispatches in parallel \citep{kim2024llmcompiler}; and Self-Discover composes reasoning structures
from modules at inference time \citep{zhou2024selfdiscover}. Their starting point is a task
description or a planner, and none of them is judged by the state a run leaves in a system.

\subsection{Compiling agent traces}
A successful trace has become the raw material of a line that turns it into an artifact later
executions can reuse. TraceCompiler mines clusters of noisy traces into a mostly deterministic
workflow \citep{tracecompiler2026}. It admits a dependency between two calls only when an argument
of the consumer holds a value that only the producer could have supplied, records the evidence for
every edge it keeps, and leaves an ambiguous relation unordered. The bindings it extracts are
classified as constants, user inputs, copied outputs, transforms or residual model decisions, and
an intent whose irreversible side effect is under-determined is refused rather than compiled. On
AppWorld, a Venmo intent goes from 34 observed API calls to 11 at run time and passes the
benchmark's state tests in 15 of 21 leave-one-out folds. The paper states what it does not measure:
call reduction is reported, the cost of compiling is not, and no net efficiency claim is made.

FlashScript turns one successful run into a neuro-symbolic program, following a set of authoring
guidelines \citep{flashscript2026}. Actions and data transformations become code, instance-specific
values become parameters, and a decision that needs model judgment becomes a typed call at a fixed
point in the program, which the guidelines place in one of three layers according to how
information moves through the workflow. Three techniques remove most of those calls: caching a
value observed in the run behind a guard, codifying a judgment as a rule, and combining independent
questions into one call. On 120 WorkBench analytics tasks the programs use about 70 times fewer
input tokens and a third of the runtime of a skill-guided agent, are correct on every repeat, and
generalize to 108 unseen instances. Their ablation matters for our setting: a program written from
the run alone generalizes markedly worse than one written from the request template, so correctness
on unseen instances comes from the template, and the run contributes tighter guards and a record of
the decisions to review.

Skill-DisCo distils parameterized control-flow subgraphs from successful traces in scenarios whose
dynamics form a finite-state machine, and compiles each into a callable, executable and verified
procedural skill \citep{skilldisco2026}. On ALFWorld and WebArena the compiled skills raise success
and reduce turns and token cost across model scales, and a compact library transfers procedures
from a stronger inducing model to a smaller executing one. Two further systems make the same trade
in narrower settings. PreAct compiles a successful run on a graphical interface into a state
machine of screen checks and actions, which it replays without per-step model calls and 8.5 to 13
times faster, handing control back to the agent when the screen does not match what the program
expects \citep{preact2026}. Neuro-symbolic skill induction lifts traces into logic-grounded
programs with explicit control flow and variable binding, so that a few traces yield skills that
generalize to unseen goals \citep{liftingtraces2026}.

These systems agree on the direction: remove from the artifact whatever code can decide, and leave
a model only where judgment is needed. They differ in four respects that matter here. The artifact
is a workflow, a program, a skill library or a state machine. The executor is an engine for the
workflow and the program, and an agent for the skills. A model call returns a value like any other,
so the code around it decides what the value does, and what the model may answer is not a closed
set of candidates that the artifact declares and can reject an answer against. The quantity
reported is calls, tokens or runtime and, in one case, the cost of an episode; none reports what it
cost to compile the artifact, which is the quantity that decides whether compiling pays. Our
position follows from those four: a graph whose nodes are typed by the space of values they
produce, control flow owned by the engine rather than by the model, a decision whose answer is
checked against candidates declared before the run, and a price that includes the compilation.

\subsection{Skills and memory}
A second line reuses experience without producing a program. Voyager stores verified executable
skills in a library that the agent retrieves while it explores \citep{wang2023voyager}, and Agent
Workflow Memory induces routines from successful web trajectories and supplies them to the agent as
text \citep{wang2024awm}. ExpeL gathers trajectories and distils them into insights that condition
the next episode \citep{zhao2024expel}, and Reflexion keeps verbal reflections between attempts
\citep{shinn2023reflexion}. Memory systems retain information rather than procedure: MemGPT treats
the context window as main memory and lets the model page facts in and out of an external store
through function calls, so a later session starts with what an earlier one chose to keep
\citep{packer2023memgpt}. All of them reduce the work of the next episode, and all of them keep a
model in the loop: the retrieved skill, insight, reflection or fact is an input to a language model
that still decides each step. None produces an artifact that a runtime executes on its own, and
none of them reports what a task costs per execution once the routine exists. Our compiler uses the
same raw material, successful traces, but its output is a graph whose steps are typed, so a model
is called only where the graph says it must be, and the graph is what the runner executes.

\subsection{LLM agents and their benchmarks}
The agent itself is the third line. ReAct interleaves reasoning traces with actions
\citep{yao2023react} and Toolformer teaches a model when to call an API \citep{schick2023toolformer};
the resulting behavior is flexible and pays a model call at every step. Agents have also been built
to automate one technical workflow end to end, as in data-driven modeling for engineering
applications \citep{liu2026automating}. The benchmarks that measure
this behavior are designed around the agent as the artifact. AppWorld provides an execution
environment of nine apps and 457 APIs, tasks generated from templates, and state-based tests that
also detect collateral damage \citep{trivedi2024appworld}; $\tau$-bench adds a simulated user, a
domain policy and a reliability metric over repeated trials \citep{yao2024taubench}; WebArena builds
realistic web environments for the same purpose \citep{zhou2024webarena}; SWE-bench grades patches
against the tests of real repositories \citep{jimenez2024swebench}; and AgentBench compares models
across eight environments \citep{liu2024agentbench}. They make agent capability comparable across
models and across time.

\subsection{Cheap inference and typed decisions}
Cost-aware inference usually lowers the price of the same question: a router picks between a strong
and a weak model for each request \citep{ong2025routellm}, and a cascade escalates only when a cheap
answer is not trusted \citep{frugalgpt2024}. ReWOO removes the repeated context instead: the plan is
written once and executed without feeding the observations back to the model \citep{xu2023rewoo}.
All three lower the price of the same question: the router and the cascade choose who answers it, and
ReWOO controls what the answer sees. We change the question at each node, and then choose who answers it. A decision model
answers one bounded question over supplied state and returns a structured answer instead of text;
the Decisions API we use is an instance of that interface \citep{jev2026}, and an escalation rule
can send the cases the model is unsure about to a larger one \citep{jevjudge2026}. An LLM has also
been used as an oracle that instantiates a symbolic artifact with domain knowledge
\citep{ciatto2025oracles}, which is the shape of a bounded question answered by a model. Used as the
lowering target of a workflow node, the question is fixed when the workflow is compiled, its
criteria may come from state observed at run time, and the answer selects a branch the engine
already knows. The typed decision is one form of the broader integration of language models with
knowledge-based components \citep{yang2025survey}.

In summary, the literature supplies the raw material (traces), the abstractions (skills, insights
and generated programs), the decision model, the benchmark and, recently, the first compilers from
traces to artifacts. What it does not supply is a settled answer to what the compiled artifact
should be, nor what compiling and running it costs against the agent it replaces. Our contribution is
both: the form of the artifact, with categories fixed by output space and routing owned by the
engine, and a measurement that includes the cost of compiling it.
 \section{The IR: EXECUTE, DECIDE and GENERATE}
\label{sec:ir}
\subsection{What a workflow is}
A compiled workflow is a directed graph with one entrypoint. Its nodes are the steps of the task, and
each of them does one thing: it takes values from the request or from earlier nodes, leaves its
result in a named place, and declares what may follow it. The order lives in the artifact rather
than in a prompt. A node names its continuation, either a single successor or a table that maps
route names to successors, and the engine follows it. A successor may be null, which ends the run
and returns the node's value as the output. Execution is sequential, and cycles are allowed, so a
workflow that repeats an operation over a collection is a cycle with a cursor rather than a loop
construct.

The request is immutable and its schema is declared. The state of a run starts empty and only nodes
add to it, because the shape of a value follows from the node that produced it rather than from a
declaration. A run resolves what each node needs, calls it, checks the result against the node's
declaration, stores it and moves on. The engine does this and nothing else: it makes no model call,
invents no route and prints nothing. Which model answers, how an operation is implemented and how
many steps a run may take are properties of the execution configuration rather than of the graph, so
the artifact says what must happen and not who does it. A package is the graph, the operations it
references and an example of its input, and the same graph can run against a terminal, a benchmark
or a host that replaces an operation by name.

\subsection{The three categories}
EXECUTE performs an operation that needs no semantic inference. The node names the operation and
binds its arguments, and the execution configuration resolves the name to a reviewed function that
ships with the package or to one the host provides. Operations are the workflow's only way to read
or change the system, and an EXECUTE node never calls a model. Because binding is by operation name
rather than by node, one operation can serve several nodes; every operation a graph references must
be bound before the run starts, so a missing binding fails at startup instead of halfway through a
run. An operation either returns its result directly or yields one external request whose response
completes it, and turning that response into a result belongs to the operation rather than to the
engine.

DECIDE asks one bounded question and declares what follows from its answer. A \emph{choice} question
maps at least two labels to descriptions and the answer selects one; a \emph{score} question asks
for a value on an ordered scale; a \emph{noul} question asks for the probability of a statement. The
answer is a structured object with the chosen value and, when the provider reports them,
probabilities and a confidence \citep{jev2026}. The engine, never the model, selects the successor: a
choice may route on its label, any question may branch on bands over its answer or its confidence,
or the node may declare a single successor and leave the answer as data for a later node. A label
outside the declared criteria is a runtime failure, which is what makes the answer checkable.
Criteria may also come from state, which is how a node enumerates a set that exists only at run
time, such as the genres the world holds; the validator then cannot compare labels with routes, and
the engine rejects an answer that has no route.

GENERATE produces a value whose space is not enumerated. It declares a template and an output
schema, and it performs the open-ended part of a task: drafting text, extracting a name or an
address, summarizing an observation. A GENERATE node may declare a route, but a routed node's value
selects its route, so a routed GENERATE could only return a route name, which is a bounded decision
and belongs in DECIDE. Routing on generated text is therefore not something the IR expresses, and
that is deliberate: free text never chooses the next node. Table~\ref{tab:categories} summarizes the
three categories by what they can produce and by what can check them.

\begin{table}[tbp]
\centering
\footnotesize
\setlength{\tabcolsep}{4pt}
\begin{tabularx}{\textwidth}{@{}l Y{1.35} Y{0.8} l Y{0.85}@{}}
\toprule
Category & What it does & Output space & Executed by & Checked against \\
\midrule
EXECUTE & names an operation, or computes over values at hand & the operation's return type & code & the system it changes \\
DECIDE & answers one bounded question over supplied state & a declared label, a scale value, or a probability & a decision model & the declared criteria \\
GENERATE & performs the open-ended part of a task & any value matching a schema & a generative model & its schema only \\
\bottomrule
\end{tabularx}
\caption{The three node categories. The partition is by output space, and the rows differ in what can
execute the node and what can verify its result.}
\label{tab:categories}
\end{table}

\subsection{Completeness of the partition}
The three categories are not a taxonomy of agent behavior. They partition the steps of a trace by the
space their value ranges across, and the partition is complete: a step either computes a value from
values at hand, or chooses among a set of candidates that the system can enumerate, or produces a
value whose space is not enumerated. The categories differ in what can execute them, namely code, a
typed decision model or a generative model, and in what can check them, namely an operation's
postcondition, the declared criteria, or only a schema. Those two differences are what the cost of a
compiled workflow depends on, which is why the partition is the design commitment of this paper
rather than a naming convention.

Merging any two categories loses one of the differences. If DECIDE were an EXECUTE node, the
operation would have to call a model, which hides the only model call in the workflow inside opaque
code and makes its cost invisible to the artifact; that is the arrangement the paper is trying to
replace. If DECIDE were a GENERATE node, free text would choose a branch, the answer could not be
validated against a closed set, and routing would move from the engine to the model. Splitting
further would not add a difference that the artifact can express: whether a decision is answered by
a large model, a small one or a person, and whether it is phrased as a label, a score or a
probability, are properties of the execution configuration rather than of the graph. Keeping them out
of the artifact is what allows the same workflow to run with a different decision backend without
editing it, which is what the experiment does when it prices a decision.

A step can also move between categories as an artifact improves, and the movement is the point of
compiling. A GENERATE that extracts a year from a sentence becomes EXECUTE once the patterns are
known, as happens in one of our families; a DECIDE becomes EXECUTE when the system can answer the
question itself, for instance when a filter accepts a genre as an argument instead of returning all
artists. The category is therefore a statement about the current artifact rather than about the
task: it records what still needs inference after the traces have been read.

\subsection{The alarm workflow}

\begin{figure}[tbp]
\centering
\resizebox{\textwidth}{!}{\begin{tikzpicture}[
  font=\sffamily\scriptsize,
  exec/.style={rectangle, rounded corners=1.5pt, draw=black!45, fill=execfill,
    minimum height=0.9cm, text width=1.0cm, align=center, inner sep=1.5pt,
    line width=0.4pt},
  dec/.style={diamond, aspect=1.9, draw=black!45, fill=decidefill,
    minimum width=2.15cm, minimum height=1.05cm, align=center, inner sep=0.5pt,
    line width=0.4pt},
  gen/.style={trapezium, trapezium left angle=70, trapezium right angle=110,
    trapezium stretches=true, draw=black!45, fill=genfill,
    minimum height=0.9cm, text width=1.05cm, align=center, inner sep=1.5pt,
    line width=0.4pt},
  flow/.style={-{Stealth[length=3.6pt,width=3pt]}, line width=0.45pt, draw=black!70},
  lab/.style={font=\sffamily\scriptsize, inner sep=1pt, fill=white},
  note/.style={font=\sffamily\tiny, text=black!55, align=center}
]
\node[exec] (list) at (1.05,0) {list\\alarms};
\node[dec] (which) at (3.2,0) {which alarm?};
\node[dec] (dir) at (6.45,0) {direction?};
\node[exec] (set) at (8.85,0) {set\\alarm};
\node[exec, text width=1.15cm] (dis) at (10.7,0) {disable\\others};
\node[exec, text width=1.15cm] (done) at (12.3,0) {complete};
\node[gen] (report) at (3.2,-2.0) {report\\missing};

\draw[flow] (list) -- (which);
\draw[flow] (which) -- node[lab, above=1pt] {an alarm} (dir);
\draw[flow] (which) -- node[lab, left=1pt, pos=0.5] {none of these} (report);
\draw[flow] (dir) -- (set);
\draw[flow] (set) -- (dis);
\draw[flow] (dis) -- (done);

\begin{scope}[yshift=-1.35cm, xshift=7.6cm]
\node[exec, minimum width=0.55cm, minimum height=0.34cm, text width=0.3cm, inner sep=0pt] (l1) at (0,0) {};
\node[right=1.4mm of l1, font=\sffamily\scriptsize, anchor=west] {EXECUTE, code};
\node[dec, minimum width=0.6cm, minimum height=0.34cm, text width=0.2cm, inner sep=0pt] (l2) at (0,-0.55) {};
\node[right=1.4mm of l2, font=\sffamily\scriptsize, anchor=west] {DECIDE, decision model};
\node[gen, minimum width=0.6cm, minimum height=0.34cm, text width=0.25cm, inner sep=0pt] (l3) at (0,-1.1) {};
\node[right=1.4mm of l3, font=\sffamily\scriptsize, anchor=west] {GENERATE, generative model};
\end{scope}
\end{tikzpicture}}
\caption{A workflow for the alarm request of Section~\ref{sec:intro}, simplified. Boxes are EXECUTE,
diamonds DECIDE and the parallelogram GENERATE; edge labels are routes. Any path calls a model at
most twice. The compiled package for this request resolves it with the first two categories only.}
\label{fig:example}
\end{figure}
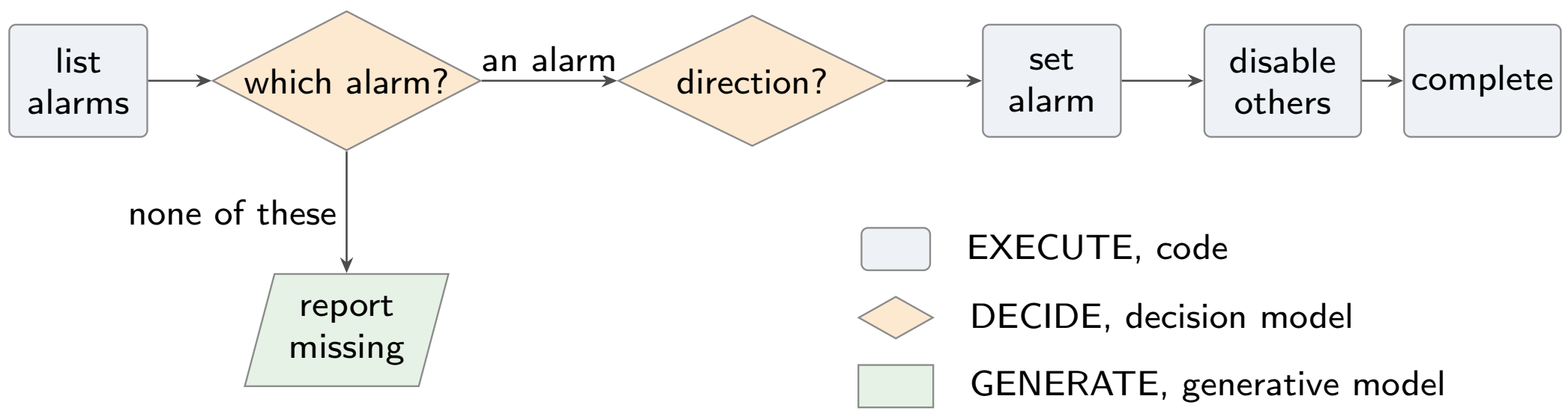

In order to illustrate the IR, Figure~\ref{fig:example} shows a workflow for the alarm request of Section~\ref{sec:intro}, written in
the IR and simplified to the steps that matter here.

The request arrives with the task identifier, the instruction and the time. The first node lists the
alarms of the account. The graph then does what the sentence cannot: it asks which of the alarms the
user means, with the labels and descriptions of those alarms as the options and one more for the case
where none of them is the one, so the question is bounded by state the run has just read. If the
answer names an alarm, the run continues on that route; if it names none of them, it takes the only
edge to a GENERATE node, which writes the explanation and ends the run. The second decision asks whether the alarm moves earlier or later and
declares a single successor: its answer is not a route but a value, which the operation that sets the
alarm reads. The remaining nodes set the alarm, disable the others and mark the task complete.

The figure shows the three categories doing their job in one workflow. Every step belongs to one of the three
categories, and the category fixes what can run it: four steps are code, two are bounded questions a
decision model can answer, and one must be generated by an LLM. The model calls are visible in the artifact
and can be counted before the run, at most two on any path. The routes are part of the graph, so the
path for a request that names no alarm is a declared edge rather than something a model improvises.
And every value in the normal path comes either from an operation or from a decision whose candidates
the node declares.
 \section{Compilation}
\label{sec:compilation}
Section~\ref{sec:ir} defined the graph IR specification; this section describes the compiler agent that produces one.

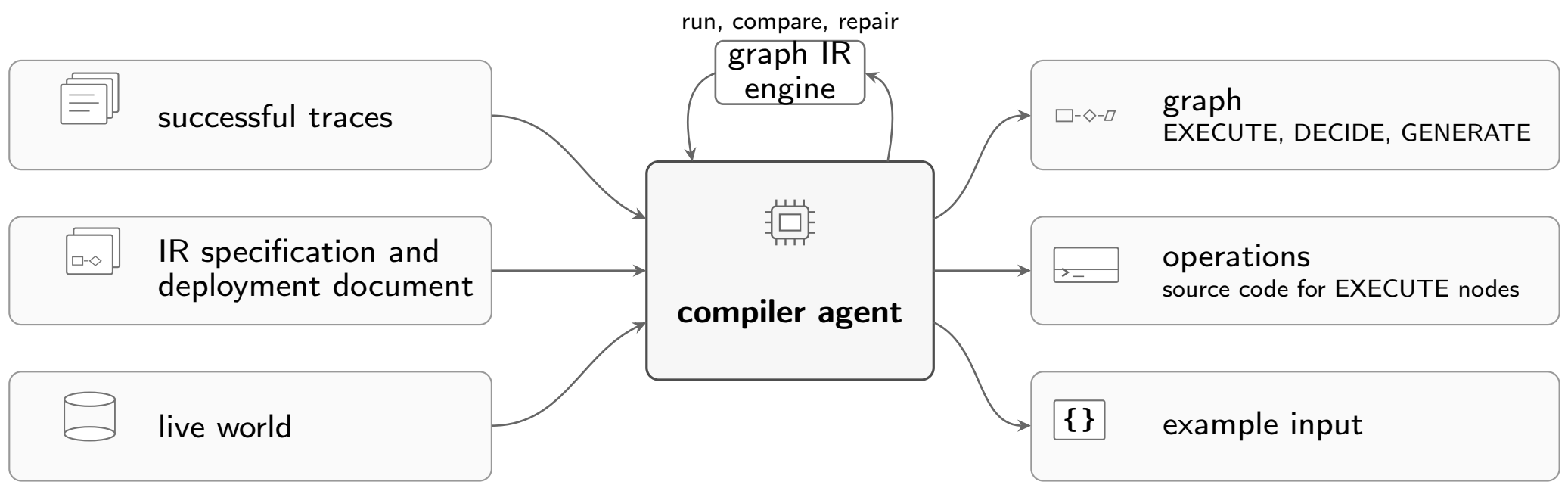
\begin{figure}[tbp]
\centering
\resizebox{\textwidth}{!}{\begin{tikzpicture}[font=\sffamily\scriptsize,
  chip/.style={rectangle, rounded corners=3pt, draw=black!40, fill=black!2, line width=0.4pt},
  comp/.style={rectangle, rounded corners=3pt, draw=black!70, fill=black!3, line width=0.6pt},
  eng/.style={rectangle, rounded corners=2pt, draw=black!55, fill=white, line width=0.5pt},
  flow/.style={-{Stealth[length=3.4pt,width=3pt]}, line width=0.5pt, draw=black!60},
  page/.style={draw=black!55, fill=white, line width=0.4pt, rounded corners=0.5pt},
  line/.style={draw=black!55, line width=0.3pt},
  lab/.style={anchor=west, align=left, font=\sffamily\scriptsize}
]
\draw[chip] (0,0.88) rectangle (4.2,1.83);
\draw[chip] (0,-0.47) rectangle (4.2,0.47);
\draw[chip] (0,-1.83) rectangle (4.2,-0.88);
\draw[page] (0.55,1.38) rectangle ++(0.40,0.34);
\draw[page] (0.50,1.33) rectangle ++(0.40,0.34);
\draw[page] (0.45,1.28) rectangle ++(0.40,0.34);
\draw[line] (0.52,1.53) -- (0.74,1.53);
\draw[line] (0.52,1.46) -- (0.78,1.46);
\draw[line] (0.52,1.39) -- (0.70,1.39);
\node[lab] at (1.15,1.35) {successful traces};
\draw[page] (0.56,0.03) rectangle ++(0.40,0.34);
\draw[page] (0.50,-0.03) rectangle ++(0.40,0.34);
\draw[execfill, draw=black!55, line width=0.25pt] (0.55,0.05) rectangle ++(0.09,0.07);
\draw[black!55, line width=0.25pt] (0.65,0.085) -- (0.69,0.085);
\draw[decidefill, draw=black!55, line width=0.25pt] (0.70,0.085) -- (0.75,0.125) -- (0.80,0.085) -- (0.75,0.045) -- cycle;
\node[lab] at (1.15,0) {IR specification and\\[-1pt]deployment document};
\draw[page] (0.70,-1.12) ellipse (0.22 and 0.06);
\draw[black!55, line width=0.35pt] (0.48,-1.42) -- (0.48,-1.12);
\draw[black!55, line width=0.35pt] (0.92,-1.42) -- (0.92,-1.12);
\draw[page] (0.70,-1.42) ellipse (0.22 and 0.06);
\node[lab] at (1.15,-1.35) {live world};

\draw[comp] (5.55,-0.95) rectangle (8.05,0.95);
\draw[black!60, line width=0.45pt, fill=black!4, rounded corners=1pt] (6.64,0.28) rectangle (6.96,0.56);
\draw[black!60, line width=0.4pt, fill=white] (6.71,0.35) rectangle (6.89,0.49);
\foreach \x in {6.68,6.76,6.84,6.92}{\draw[black!60, line width=0.45pt] (\x,0.56) -- (\x,0.62);}
\foreach \x in {6.68,6.76,6.84,6.92}{\draw[black!60, line width=0.45pt] (\x,0.28) -- (\x,0.22);}
\foreach \y in {0.33,0.42,0.51}{\draw[black!60, line width=0.45pt] (6.64,\y) -- (6.58,\y);}
\foreach \y in {0.33,0.42,0.51}{\draw[black!60, line width=0.45pt] (6.96,\y) -- (7.02,\y);}
\node[font=\sffamily\scriptsize\bfseries] at (6.80,-0.38) {compiler agent};

\draw[eng] (6.15,1.45) rectangle (7.45,2.00);
\node[font=\sffamily\scriptsize, align=center] at (6.80,1.725) {graph IR\\[-1pt]engine};
\draw[flow] (6.15,1.72) to[out=200,in=100] (5.95,0.95);
\draw[flow] (7.65,0.95) to[out=80,in=-20] (7.45,1.72);
\node[font=\sffamily\tiny] at (6.80,2.16) {run, compare, repair};

\draw[flow] (4.2,1.35) to[out=0,in=155] (5.55,0.45);
\draw[flow] (4.2,0) to[out=0,in=180] (5.55,0);
\draw[flow] (4.2,-1.35) to[out=0,in=205] (5.55,-0.45);
\draw[flow] (8.05,0.45) to[out=25,in=180] (8.9,1.35);
\draw[flow] (8.05,0) to[out=0,in=180] (8.9,0);
\draw[flow] (8.05,-0.45) to[out=-25,in=180] (8.9,-1.35);

\draw[chip] (8.9,0.88) rectangle (13.5,1.83);
\draw[chip] (8.9,-0.47) rectangle (13.5,0.47);
\draw[chip] (8.9,-1.83) rectangle (13.5,-0.88);
\draw[execfill, draw=black!55, line width=0.35pt] (9.12,1.30) rectangle ++(0.14,0.10);
\draw[black!55, line width=0.3pt] (9.28,1.35) -- (9.33,1.35);
\draw[decidefill, draw=black!55, line width=0.35pt] (9.36,1.35) -- (9.41,1.40) -- (9.46,1.35) -- (9.41,1.30) -- cycle;
\draw[black!55, line width=0.3pt] (9.48,1.35) -- (9.53,1.35);
\draw[genfill, draw=black!55, line width=0.35pt] (9.54,1.31) -- (9.60,1.31) -- (9.63,1.39) -- (9.57,1.39) -- cycle;
\node[lab] at (9.9,1.35) {graph\\[-2pt]{\tiny EXECUTE, DECIDE, GENERATE}};
\draw[page] (9.10,-0.10) rectangle ++(0.56,0.30);
\draw[black!55, line width=0.3pt] (9.10,0.01) -- (9.66,0.01);
\draw[black!60, line width=0.45pt] (9.17,-0.05) -- (9.22,-0.015) -- (9.17,0.02);
\draw[black!60, line width=0.3pt] (9.26,-0.05) -- (9.36,-0.05);
\node[lab] at (9.9,0) {operations\\[-2pt]{\tiny source code for EXECUTE nodes}};
\draw[page] (9.10,-1.47) rectangle ++(0.44,0.34);
\node[font=\sffamily\bfseries\tiny] at (9.32,-1.30) {\{\,\}};
\node[lab] at (9.9,-1.35) {example input};
\end{tikzpicture}
}
\caption{The compiler and its inputs. It reads the traces, the IR specification, the deployment document
and a world of traced instances, and writes a graph, its operations and an example input. It checks
the package in the engine against the world and repairs it until the state matches the traces.}
\label{fig:compiler}
\end{figure}

\subsection{The workspace and the package}

The compiler receives three inputs.
\begin{itemize}
  \item \textbf{The traces}, one per traced instance, recording the tracing agent's reasoning, each
  block of code it sent, each output it received and the token counts of every turn.
  \item \textbf{The written specification, in two documents.} The IR specification is prose and the
  schema the engine enforces, so what the compiler validates against is what the engine will check.
  The deployment document says what the runner supplies, the task's identifier, the instruction in
  the user's words, the time of the request and the supervisor's name and contact details, and that
  the workflow must tell the system it is done, because the runner ends the process and not the task.
  \item \textbf{A live world} that holds the traced instances: the benchmark's databases, API
  documentation and datasets, and the task directories of the traced tasks.
\end{itemize}

The compiler is given no task-specific instructions. The instructions describe how to read a trace,
what to write and how to check it; the deployment document describes the runner; neither names an
app, a field or a fact of the task it has to reproduce (an alarm, an artist, a song) nor describes the
instances on which the package will be judged. The traces are the only description of
the task it receives. It has the ordinary file and shell tools inside a sandbox, and it does not have
the benchmark's tool for calling the apps: the documented way to reach the world is the package it
writes.

The compiler's workspace holds the world of the traced instances only. The directories, reference
actions and grading material of the held-out instances are absent, because a compiler that could read
the instances it will be judged on could hard-code what they expect, and the comparison would then
measure the compiler's access instead of its output.

The compiler writes a package of three parts: the graph, the operations that its EXECUTE nodes call
and an example of the input. The graph is validated against the schema the engine enforces, which
rejects a malformed package before anything runs, but validation only checks the form of the graph.
The compiler therefore runs the package in the engine against the live world, compares the state
changes with the traces and repairs what does not match, until a run reaches its terminal node with
the state the traces produced. A package whose operations answer from data they carry reproduces the
trace by construction and has checked nothing, which the compiler's instructions state. The workspace
and the world are rebuilt for every compile, so two compiles differ because of the compiler alone.

\subsection{Deployment details}

The method is generic, but its output is a package that a host calls, and a trace fixes what the
workflow does, not how it is called. Three points of that interface were left open, and each of them
cost a compile before we wrote it down. The statements live in the IR specification and in the
deployment document, which are inputs to the compile and say nothing about the task. The document is
short: it names the five fields of the task record that the runner supplies, with the supervisor's
first name, last name, email and phone number, says that the input declaration is filled from them,
and states that the workflow must tell the system it is done. A real deployment would settle them
before the compilation, and the artifact would be built against them. AppWorld does not, so we state
them there to fix the evaluation, and every result in this paper was produced under the final version
of each.

\paragraph{An operation must reach the system instead of simulating it}
In earlier experiments, we saw that one compile rebuilt the artists it had seen in the traces, ran its 
workflow against that copy and reported the workflow as verified, because the copy agreed with itself. 
The statement that resolved this is in the IR specification: operations are the workflow's only way to 
read or change the system, so a value they return has to come from that system.

\paragraph{The input is what the caller knows}
Three compiles of the same traces declared three different input shapes, and all three were consistent
with what the traces show: a trace records the values the agent was given and the values it fetched
side by side, and it cannot say which of them the artifact should be handed. The graphs were equally
valid and not interchangeable, because a runner fills one declaration and refuses the others. We
therefore fixed the declaration for the evaluation: the deployment document states that the runner
supplies the task record and nothing else, that is, the task's identifier, the instruction, the time
of the request and the supervisor's details, and a package that asks for anything else is refused
before it runs. Each of the three shapes would run from a caller that supplied it, so the choice is a
property of the deployment, not of the compilation.

\paragraph{The runner ends the process, not the task}
The third statement says that the workflow must tell the system it is done, as the traces did, and
that a terminal node is not a completion. In all traces of this study the last action is a call that
tells the supervisor the task is complete, recorded like any other step, and the world records a task
as complete only when it receives that call. One compile reached its terminal node, re-read the
account and confirmed the state the traces had produced, and made no such call, so every change was
right and the task was incomplete. The traces state this to the agent they record, as a way to finish
its session, and not to the artifact as part of its interface. The deployment document states it as
one: the workflow must tell the system that it is done itself, and the runner ends the process at the
terminal node without doing it on the workflow's behalf. In another deployment the same requirement
appears as the action that closes the work, such as resolving a ticket or notifying the requester, and
it is no less part of the task.
 \section{Experimental Setup}
\label{sec:setup}

This section describes the benchmark, the split, the agents and the cost accounting.

\subsection{AppWorld}
AppWorld is an execution environment of nine day-to-day apps reachable through 457 APIs, populated
with the digital lives of about a hundred users, together with a benchmark of tasks evaluated by the
state of the database and by the task's own tests, which also detect collateral damage
\citep{trivedi2024appworld}. Each task comes from a generator that fixes the request and varies the
world: the instruction is the same template with different values, and the data it refers to changes
from instance to instance. A generator therefore produces what this paper needs, many executions of
one procedure under different data, which is the setting in which a fixed script breaks and a few
traces still describe the workflow. It also keeps tracing, compilation and evaluation inside one
benchmark whose tasks are realistic and whose evaluation is the state of the system.

We use three generators, chosen because they are multi-step, they change state, and their
instructions differ in how much of the request is a name the system can enumerate:
\begin{itemize}
  \item \textbf{Artists} (\code{07b42fd}). \emph{Follow all the classical artists on Spotify that have
  at least 22 followers.} The agent logs in as the supervisor, lists the artists of the genre, keeps
  those above the threshold, follows each one and reports completion. The genre and the follower
  threshold vary between instances.
  \item \textbf{Alarms} (\code{302c169}). \emph{Move my wake-up phone alarm 40 minutes earlier and
  disable the rest.} The agent logs in, lists the alarms, identifies the wake-up one, shifts its
  time, disables the others and reports completion. The alarm type (wake-up or go-to-sleep), the
  direction (earlier or later) and the amount (20, 40 or 50 minutes, or 1, 1.5 or 2 hours) vary
  between instances.
  \item \textbf{Songs} (\code{ce359b5}). \emph{Remove all songs from my Spotify library and playlists
  that were released before 2021.} The agent fetches the library and the playlists, reads the release
  date of each song, removes the ones that satisfy the comparison and reports completion. The
  comparison (before, after, in or before, in or after) and the year vary between instances.
\end{itemize}

\subsection{Instances and split}
We generate with the benchmark's own generators, using up to six seeded processes per family, and we deduplicate by a signature over the instruction, the supervisor, the time of the request and the contents of the databases. The result is 76 distinct instances (25, 24 and 27) covering 8, 6 and 6 distinct instructions. Ten instances per family are held out and the remaining 46 are traced, 15, 14 and 17 respectively. The split is constrained: one instance of every instruction goes to the traced side first, so every held-out instruction also appears in at least one trace. What the held-out set withholds is the world, that is, databases, users, identifiers and dates, and not
the request, so what a held-out instance adds is a world the compiler has not seen. Each generator
process yields about six tasks and then exhausts its search space, and the instruction space of the
three families is 8, 6 and 6 points wide, so further seeds add worlds at the same requests, which is
what the split needs.
\subsection{Agents and models}
The tracing agent is \code{pi} with DeepSeek V4.1 Flash at high thinking, given the benchmark's own
instructions ported to the chat channel and one tool that executes Python in the task's apps;
AppWorld exposes the apps as a Python library, so that tool is how any agent reaches them, and a run
took between 6 and 18 tool calls. The compiler is the same harness and the same model without that tool, given read, bash, edit and write inside a sandbox that holds the workspace, the engine and a writable run directory. The runtime answers DECIDE with Jev through the Decisions API, using the preview snapshot of the 1.14 generation \citep{jev2026}, and GENERATE with DeepSeek V4.1 Flash through a chat endpoint; none of the three main packages
contains a GENERATE node, and the coverage compiles that do are reported in
Section~\ref{sec:results}. The engine's step budget is 1000, and the largest run we observed took 130 nodes.

\subsection{Cost accounting}
Every run records the tokens it reported, input, cached input and output, and we price them afterwards
from a table read on 21 October 2026 at peak rates. Each cost appears twice. The first prices cached
input at the cache rate and is what the run paid; the second prices every input token at the miss rate,
which is the more expensive figure and the one usually reported in the literature. The difference is
not small: prefix caching moves the agent by a factor of eleven and the all-generate baseline by five,
because both reread context that a workflow does not carry. We also report input tokens per run, which
no price list changes, and the median wall clock per run.
 \section{Results}
\label{sec:results}

This section reports what compilation produces and what it costs.
Section~\ref{sec:comparison} compares the compiled workflows with the tracing agent on the held-out
instances, Section~\ref{sec:lowers} reads the three compiled graphs and asks which steps lowered,
Section~\ref{sec:coverage} how many traces that takes, Section~\ref{sec:stability} whether a
recompile gives the same workflow, and Section~\ref{sec:ablations} prices the two changes that the
comparison makes at once.

\subsection{The main comparison}
\label{sec:comparison}
On the 30 held-out instances, 10 per family, with every trace given to the compiler, the compiled
workflows solve 30/30 and the agent solves 29/30. The difference in cost is larger than the
difference in success. The median run costs \$0.000023 against \$0.002276, 98 times less; the means
are \$0.000019 and \$0.002644, a ratio of 139, because ten of the thirty workflow runs cost nothing.
The median wall clock is 0.6 seconds against 46, about 76 times faster, and the workflow makes 1.0
model call per run against 10.1.

Two readings of the comparison are possible and only one of them is ours. The first is that the
workflow is a better agent. We do not claim it, and ten instances per family cannot establish it: a
10/10 result is consistent with a true rate as low as 0.74, and 30/30 with 0.90, which is why the
paper reports cost at comparable success, 30/30 against 29/30, and not a success improvement. The second reading is that the
workflow reproduces the behavior the traces recorded, on worlds it has never seen, at a small
fraction of the cost. That is what the comparison shows, and the single agent failure is a reminder
that it is between one agent run and one workflow run per instance, not between two curated datasets.

\begin{figure}[tbp]
\centering
\begin{tikzpicture}
\begin{groupplot}[
  group style={group size=3 by 1, horizontal sep=0.6cm},
  width=0.28\textwidth, height=4.2cm,
  xmin=0, xmax=60, ymin=0, ymax=0.15,
  ytick={0,0.05,0.1,0.15}, yticklabels={0,0.05,0.1,0.15},
  xtick={0,20,40,60},
  xlabel={executions},
  tick label style={font=\footnotesize},
  label style={font=\footnotesize},
  title style={font=\small},
  grid=major, grid style={black!10},
]
\nextgroupplot[title={Artists}, ylabel={accumulated cost (USD)}]
\addplot[dash pattern=on 5pt off 3pt, black] table[x index=0, y index=1] {figures/break-even-artists.dat};
\addplot[line width=1pt, black] table[x index=0, y index=2] {figures/break-even-artists.dat};
\draw[dashed, black!45] (axis cs:50,0) -- (axis cs:50,0.15);
\nextgroupplot[title={Alarms}, yticklabels={}]
\addplot[dash pattern=on 5pt off 3pt, black] table[x index=0, y index=1] {figures/break-even-alarms.dat};
\addplot[line width=1pt, black] table[x index=0, y index=2] {figures/break-even-alarms.dat};
\draw[dashed, black!45] (axis cs:41,0) -- (axis cs:41,0.15);
\nextgroupplot[title={Songs}, yticklabels={}]
\addplot[dash pattern=on 5pt off 3pt, black] table[x index=0, y index=1] {figures/break-even-songs.dat};
\addplot[line width=1pt, black] table[x index=0, y index=2] {figures/break-even-songs.dat};
\draw[dashed, black!45] (axis cs:31,0) -- (axis cs:31,0.15);
\end{groupplot}
\node[anchor=north] at ($(group c2r1.south) + (0,-0.9)$) {\begin{tikzpicture}[baseline=-0.5ex]
    \draw[dash pattern=on 5pt off 3pt, black] (0,0) -- (0.55,0);
    \node[right, font=\small] at (0.55,0) {agent};
    \draw[line width=1pt, black] (1.9,0) -- (2.45,0);
    \node[right, font=\small] at (2.45,0) {compiled};
  \end{tikzpicture}};
\end{tikzpicture}
\caption{Accumulated cost against executions of the same request, one panel per family. The dashed line is
the agent, the solid line the compiled workflow, which pays for the traces and the compile once and then hardly
grows; the vertical line marks the break-even. Below it the agent is cheaper.}
\label{fig:break-even}
\end{figure}
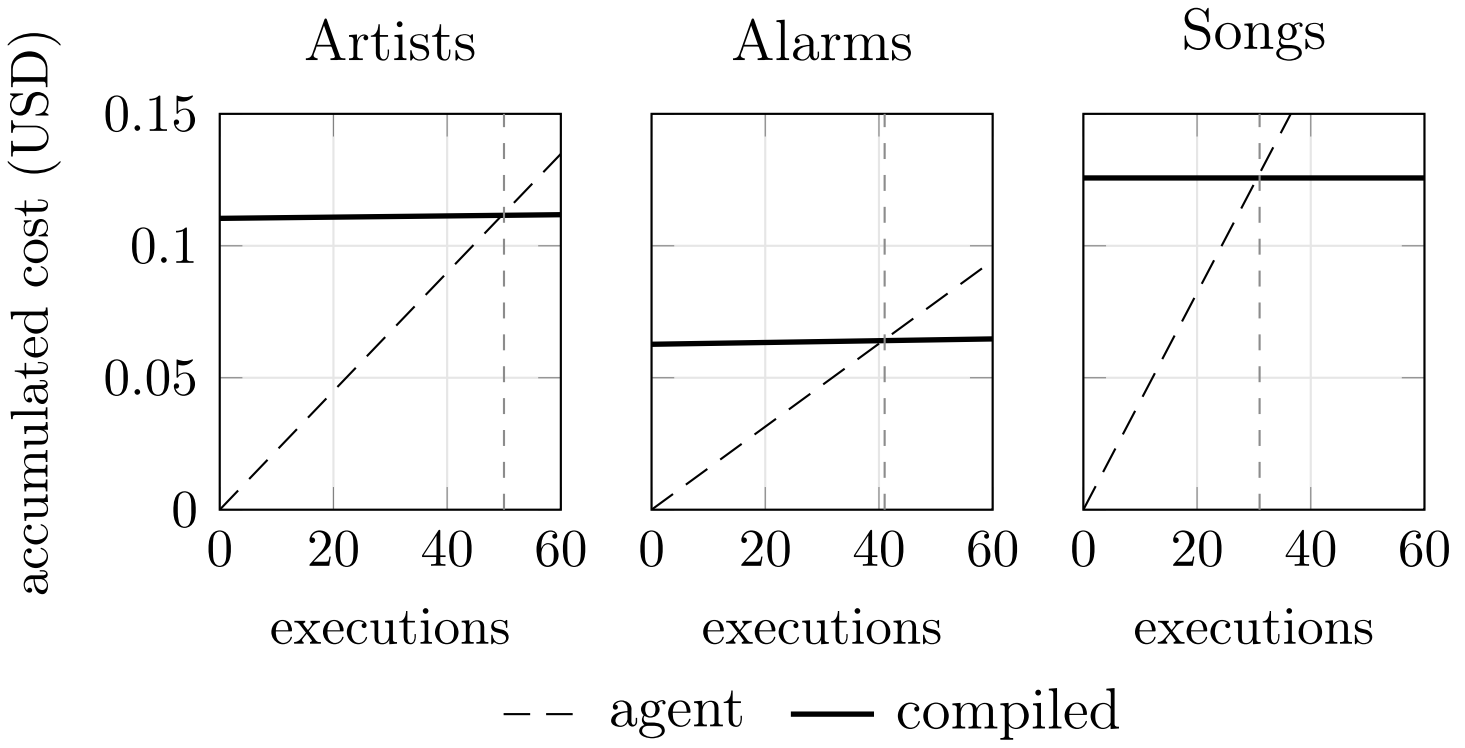

The comparison is a decision about repetition, which Figure~\ref{fig:break-even} makes explicit. Each
panel accumulates the cost of serving the same request over an increasing number of executions: the
agent's line starts at the origin and rises at its per-run cost, and the compiled line starts at what
the traces and the compilation cost and rises at the workflow's own. The slopes differ by two orders of magnitude,
so the compiled line crosses the agent's after a few dozen executions and stays below it afterwards;
the flatness of that line is the result, because once the artifact is paid for a run adds almost nothing.
The crossing is 50 executions for Artists, 41 for Alarms and 31 for Songs, the break-even in
Table~\ref{tab:main}: below it the agent is cheaper because the artifact has not been repaid, and
above it every further execution is almost free.

\begin{table}[tbp]
\centering
\small
\begin{tabular}{@{}llcrrrc@{}}
\toprule
Family & Arm & Solved & \$/run & s/run & Calls & Break-even \\
\midrule
Artists & IR    & 10/10 & 0.000023 & 0.5 & 1.0  & 50 \\
               & agent & 10/10 & 0.002247 & 46  & 9.3  & --- \\
Alarms & IR    & 10/10 & 0.000034 & 0.6 & 2.0  & 41 \\
               & agent & 9/10  & 0.001573 & 38  & 7.7  & --- \\
Songs & IR    & 10/10 & 0.000000 & 0.6 & 0.0  & 31 \\
               & agent & 10/10 & 0.004111 & 62  & 13.3 & --- \\
\midrule
pooled         & IR    & 30/30 & 0.000019 & 0.6 & 1.0  & --- \\
               & agent & 29/30 & 0.002644 & 46  & 10.1 & --- \\
\bottomrule
\end{tabular}
\caption{Compiled workflows against the tracing agent on the 30 held-out instances. Cost is the mean over a
family's ten runs, seconds the median, calls the mean per run; break-even is the cost of the traces
and the compile divided by the per-run saving, for the IR arm only.}
\label{tab:main}
\end{table}

\subsection{What lowers, and what does not}
\label{sec:lowers}
The number of model calls follows the instruction, not the size of the workflow. The song workflow
compiles to 13 EXECUTE nodes and no model call at all: the year and the comparison operator are
parsed from the sentence, and the rest of the package lists the library and the playlists, filters
the songs, removes them and verifies the removal. The artist workflow spends one call per run, on
the genre, whose candidates are the genre labels the world holds and whose criteria are therefore
read from state. The alarm workflow spends two, one on the alarm type and one on the direction of
the shift, both from a closed set of labels and both extracted from the sentence. The three packages
have 13, 16 and 20 nodes, so the difference between no calls and two is not a difference in length or
in the number of state changes; it is the difference between parameters whose value the sentence
determines and parameters that name an entity of the world, with the direction of the shift in
between, since it is a bounded choice over two words that the compiler answered with a decision.
Every dollar in the IR column is one of those calls, because no package calls a model outside a
decision node.

The song workflow is also the smallest of the three, and for the same reason. Its sentence carries a
value and an operator, and once both are parsed the task is a filter over a list. The alarm sentence
names an alarm by its role in the user's life, which the database stores as a label, and the artist
sentence names a genre that exists only as a field of the artists the world contains. A compiler
cannot decide that a string is a genre by computing; it can only ask among the genres that exist,
which is what a DECIDE node with criteria from state is for. Table~\ref{tab:packages} summarizes the three packages, where the number of model calls per run
equals the number of decision nodes.

\begin{figure}[tbp]
\centering
\small
\textbf{(a)} Artists, 20 nodes: 19 EXECUTE, 1 DECIDE\\[2pt]
\includegraphics[width=\textwidth]{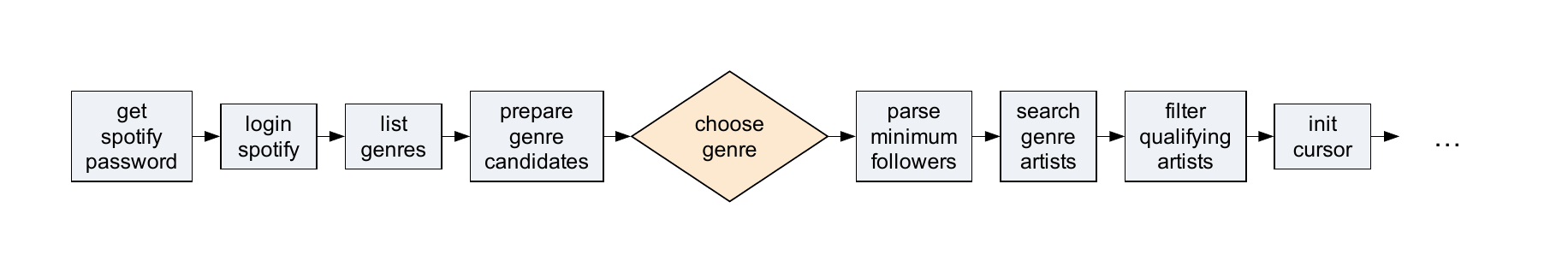}\\[-6pt]
\includegraphics[width=0.86\textwidth]{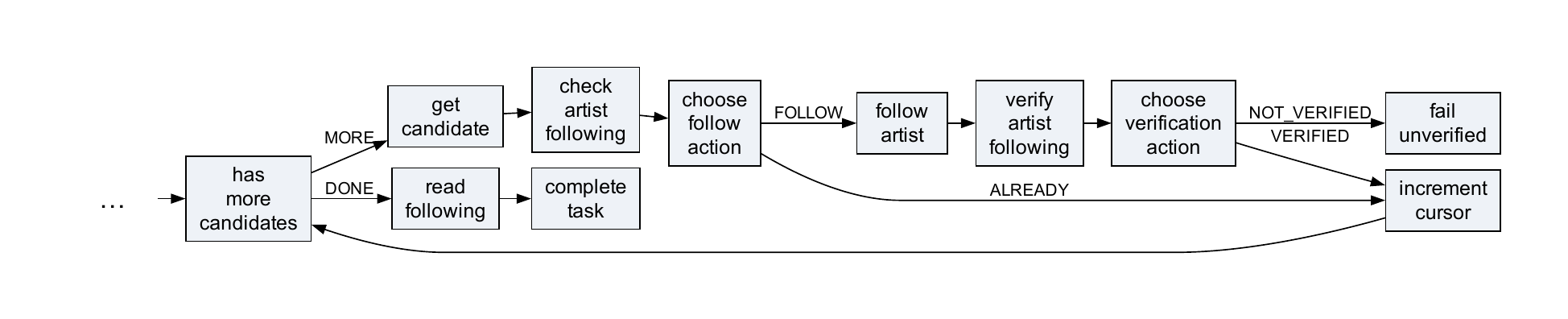}\\[6pt]
\textbf{(b)} Alarms, 16 nodes: 14 EXECUTE, 2 DECIDE\\[2pt]
\includegraphics[width=\textwidth]{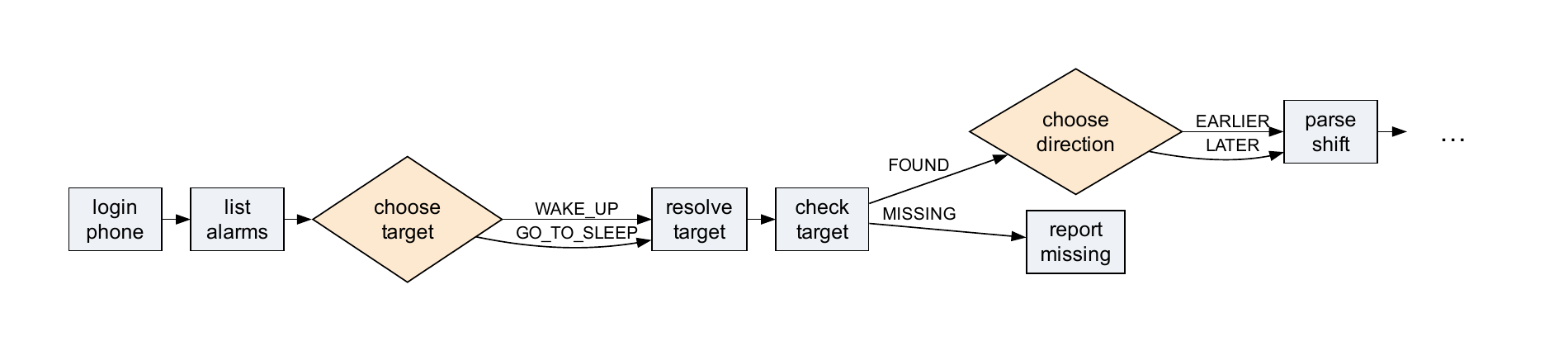}\\[-6pt]
\includegraphics[width=0.80\textwidth]{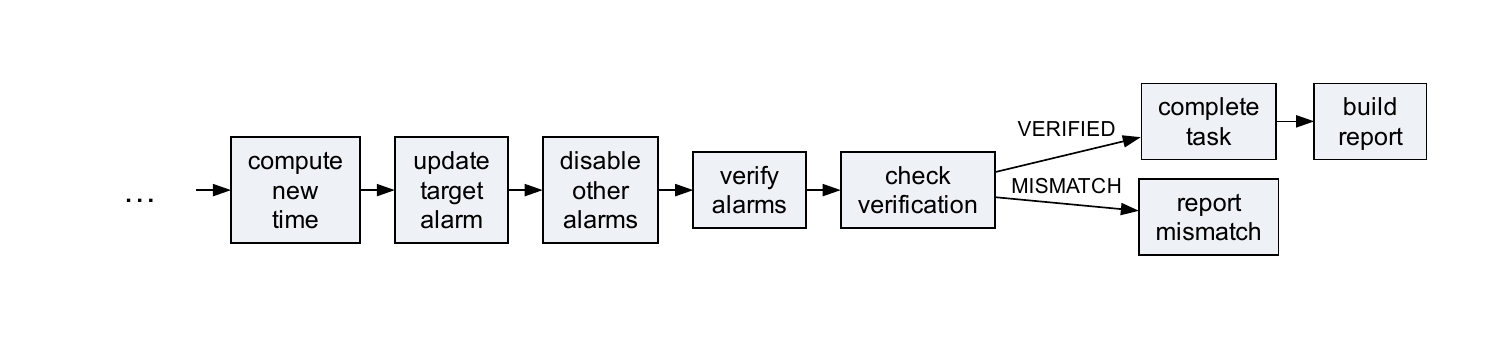}\\[6pt]
\textbf{(c)} Songs, 13 nodes: 13 EXECUTE, 0 DECIDE\\[2pt]
\includegraphics[width=0.78\textwidth]{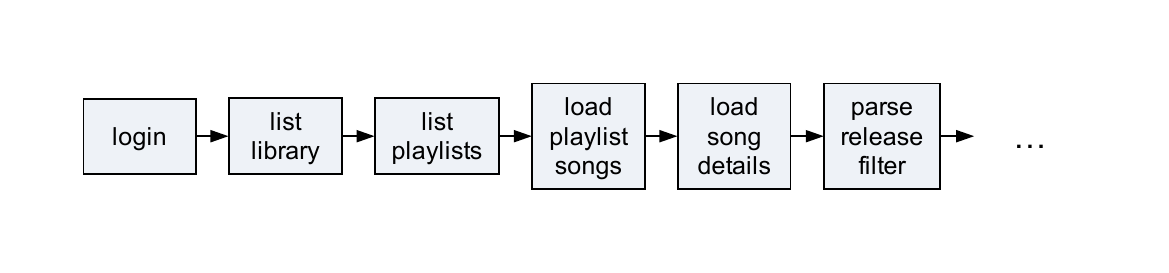}\\[-6pt]
\includegraphics[width=0.66\textwidth]{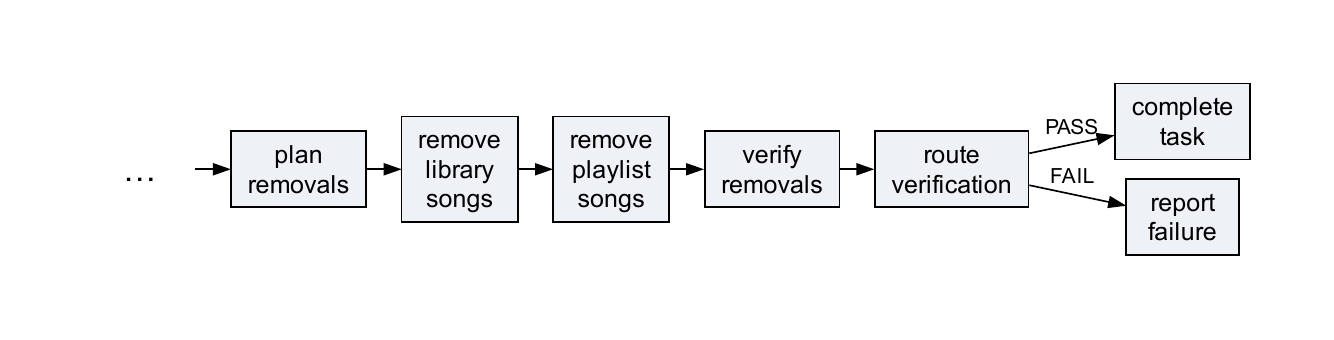}
\caption{The three compiled workflows, drawn from each package's graph, in two rows; the ellipsis marks a
continuation. Boxes are EXECUTE and diamonds DECIDE. The artist decision picks a genre from state,
the alarm decisions read labels that the next operation uses, and the song workflow has no decision.}
\label{fig:graphs}
\end{figure}

\begin{table}[tbp]
\centering
\footnotesize
\setlength{\tabcolsep}{4pt}
\begin{tabular}{@{}lrrrrrrr@{}}
\toprule
Family & Nodes & EXEC & DEC & GEN & Ops & Calls & Compile \$ \\
\midrule
Artists & 20 & 19 & 1 & 0 & 18 & 1.0 & 0.070 \\
Alarms & 16 & 14 & 2 & 0 & 14 & 2.0 & 0.041 \\
Songs & 13 & 13 & 0 & 0 & 13 & 0.0 & 0.050 \\
\bottomrule
\end{tabular}
\caption{Structure of the compiled packages. Operations are distinct operation names, not nodes, and the
last column is the cost of the compile alone; every package also contains an example input and a
report.}
\label{tab:packages}
\end{table}

The three packages differ in size by seven nodes and in model calls by two, and the family with the
fewest nodes is the one with no model call: the count of nodes is not what determines the cost. The
compile time was 4.8, 3.3 and 3.6 minutes respectively, all measured under the final specification.

\subsection{How many traces are enough}
\label{sec:coverage}
Table~\ref{tab:coverage} reports held-out success when the compiler is given the first $k$ traces of
a deterministic sample. The artist workflow is at 10/10 for every $k$, including one. The alarm
workflow falls to 4/10 at $k=1$ and the song workflow to 2/10, and both return to 10/10 at $k=3$ and
stay there for $k=5$, $k=10$ and all traces. At $k=1$ the two families still pass 59 and 54 of their 80 task tests, so the failure is partial.
The failures at $k=1$ are failures of the work rather
than of the protocol: one trace yields one instance's workflow, the compiler sees one alarm label,
one direction and one year comparison, and it encodes them as the procedure, because nothing in a
single trace says that they vary. The family whose parameter is parsed from the sentence is the one
where a single instance already shows the parameter is not a constant.

We read the curve as a property of instruction diversity and not as a claim that three traces are
enough. Three traces expose the parameter space of these generators, and after that more traces of
the same instructions add worlds instead of requests, which the flat part of the curve reflects. A
family whose instructions varied along an axis absent from the first three traces would keep failing
at ten, and the curve would not show it; the split of Section~\ref{sec:setup} deliberately keeps
every instruction on the traced side, so this experiment measures how many worlds a workflow
survives once it knows the request, not how many requests it takes to learn one.

\begin{table}[tbp]
\centering
\small
\begin{tabular}{@{}lccccc@{}}
\toprule
Family & $k=1$ & $k=3$ & $k=5$ & $k=10$ & all \\
\midrule
Artists & 10/10 & 10/10 & 10/10 & 10/10 & 10/10 \\
Alarms & 4/10  & 10/10 & 10/10 & 10/10 & 10/10 \\
Songs & 2/10  & 10/10 & 10/10 & 10/10 & 10/10 \\
\bottomrule
\end{tabular}
\caption{Held-out successes out of 10 when the compiler is given the first $k$ traces of a
deterministic sample; the $k=\text{all}$ column reuses the package of Table~\ref{tab:main}.}
\label{tab:coverage}
\end{table}

\begin{figure}[tbp]
\centering
\begin{tikzpicture}
\begin{axis}[
  width=0.94\textwidth, height=5.4cm,
  xmin=0.75, xmax=5.25, ymin=0, ymax=10.8,
  xtick={1,2,3,4,5}, xticklabels={1,3,5,10,all},
  xlabel={traces given to the compiler},
  ylabel={held-out instances solved, of 10},
  tick label style={font=\small},
  label style={font=\small},
  legend pos=south east, legend style={font=\small, draw=none, cells={anchor=west}},
  grid=major, grid style={black!12},
]
\addplot+[mark=*, thick] table[x index=0, y index=1] {figures/coverage-07b42fd.dat};
\addlegendentry{Artists}
\addplot+[mark=square*, thick] table[x index=0, y index=1] {figures/coverage-302c169.dat};
\addlegendentry{Alarms}
\addplot+[mark=triangle*, thick] table[x index=0, y index=1] {figures/coverage-ce359b5.dat};
\addlegendentry{Songs}
\end{axis}
\end{tikzpicture}
\caption{Held-out success against the number of traces given to the compiler, on a deterministic sample.
One trace is enough for the linear family, three for the others; beyond three, more traces buy
worlds, not requests.}
\label{fig:coverage}
\end{figure}
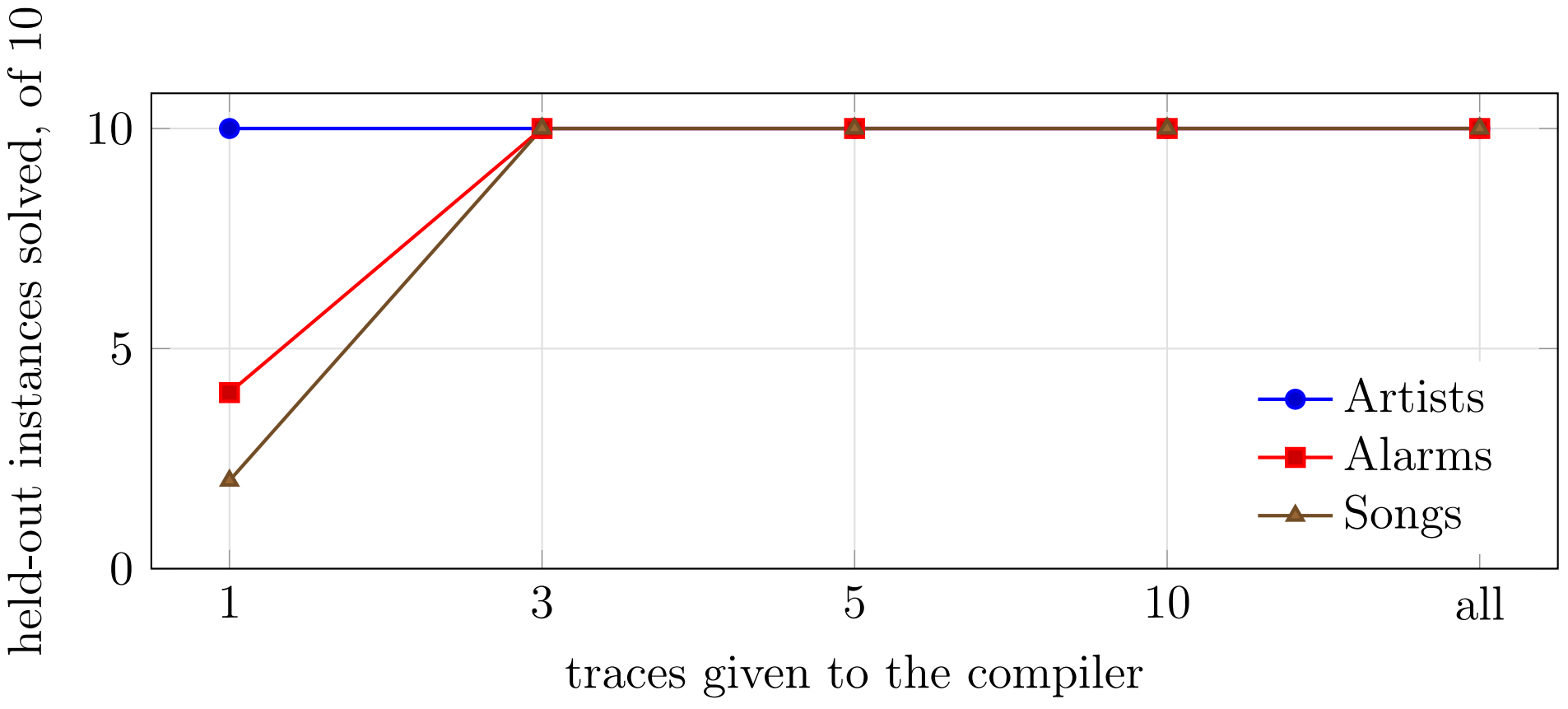

\begin{figure}[tbp]
\centering
\small
\includegraphics[width=0.86\textwidth]{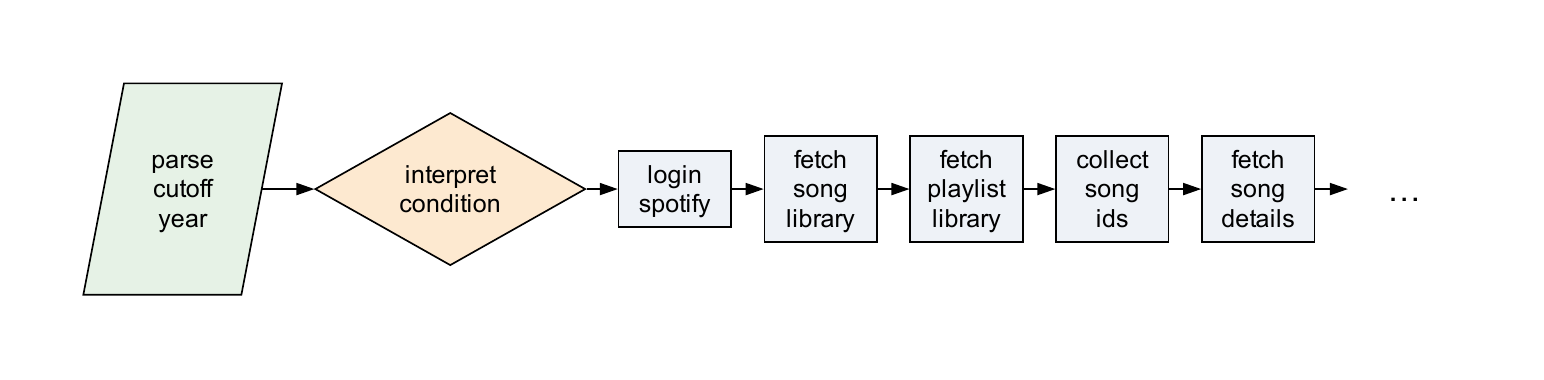}\\[-6pt]
\includegraphics[width=0.72\textwidth]{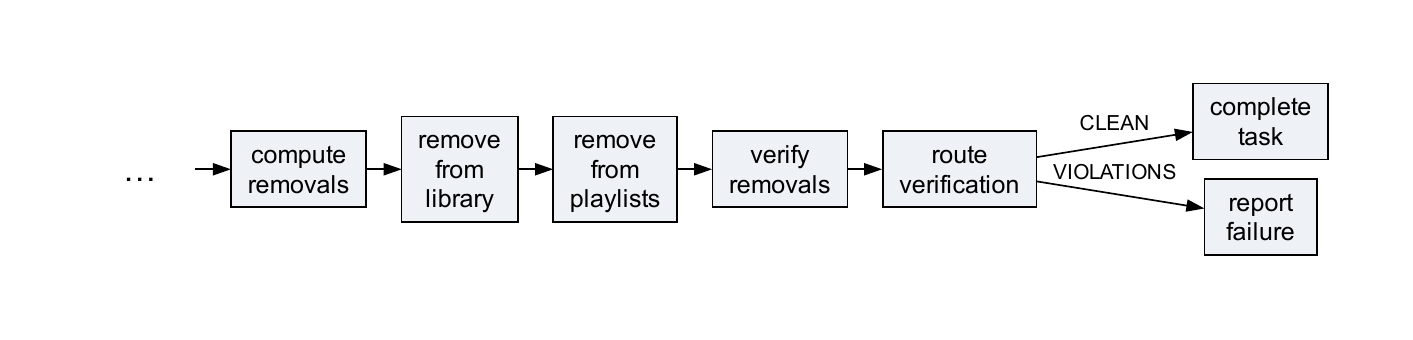}
\caption{The song workflow from three traces, 14 nodes: 12 EXECUTE, one DECIDE and one GENERATE
(parallelogram) that reads the cutoff year. The package of Figure~\ref{fig:graphs}(c) parses it in
code.}
\label{fig:generate}
\end{figure}

The packages compiled from few traces also show where the escape hatch is used. Five of the twelve
packages behind Figure~\ref{fig:coverage} contain one GENERATE node, and all five are parses of the
instruction: the cutoff year for Songs at $k=3$ and $k=10$, the alarm role and the shift in minutes
for Alarms at $k=5$ and $k=10$, and the follower threshold for Artists at $k=5$. The node is on the
main path, so it ran in all fifty held-out runs of those packages, and every run passed.
Figure~\ref{fig:generate} shows the song package at $k=3$, where the parse of the year and the
classification of the comparison are a GENERATE and a DECIDE; the all-traces package of
Figure~\ref{fig:graphs}(c) does both in code.

The one-trace packages are the other side of the same choice: there the compiler writes a parser for
the phrasing of the single trace it has, and the runs fail when the sentence is phrased differently.
The two kinds of failure are worth separating because they are fixed in different ways. On one
instance of the alarm workflow the agent moved the right alarm to the right time, disabled the right
alarms, and returned \code{'22:30'} where the task expects \code{null}: the procedure was right and
the answer convention was not, which a statement about the deployment fixes and a compile absorbs. A
compiled workflow that assumed one label, one direction and one year comparison is the other kind:
its procedure is wrong for instances whose sentence differs, and only more traces or a better
compiler fix it.

\subsection{Stability of the compiler}
\label{sec:stability}
The compiler is an LLM agent, so its output need not be a function of its input. To see whether the
measurement depends on the draw, we compiled each family three times from the same traces, the package
of Table~\ref{tab:main} and two recompiles, and all nine packages solve all ten held-out instances. Six recompiles at ceiling do not prove that
the compiler is deterministic; they show that at this resolution the result is not an artifact of one
fortunate package. The recompiles reported here are all under the final specification.

\begin{figure}[tbp]
\centering
\begin{tikzpicture}
\begin{axis}[
  width=0.66\textwidth, height=5.6cm,
  xbar stacked, bar width=0.30cm,
  ytick={1,...,9}, yticklabels={Artists 1,Artists 2,Artists 3,Alarms 1,Alarms 2,Alarms 3,Songs 1,Songs 2,Songs 3},
  y dir=reverse, ymin=0.4, ymax=9.6,
  xmin=0, xmax=21,
  xlabel={nodes}, ylabel={package},
  tick label style={font=\small}, label style={font=\small},
  legend style={font=\small, draw=none, at={(1.02,0.5)}, anchor=west},
  grid=major, grid style={black!10},
  axis x line=bottom, axis y line=left,
]
\addplot[fill=execfill, draw=black!50] table[x index=1, y index=0] {figures/composition.dat};
\addlegendentry{EXECUTE}
\addplot[fill=decidefill, draw=black!50] table[x index=2, y index=0] {figures/composition.dat};
\addlegendentry{DECIDE}
\addplot[fill=genfill, draw=black!50] table[x index=3, y index=0] {figures/composition.dat};
\addlegendentry{GENERATE}
\end{axis}
\end{tikzpicture}
\caption{The nodes of the nine compiled packages, split by category. The three compiles of a family differ
in size and composition, and all solve the ten held-out instances.}
\label{fig:composition}
\end{figure}
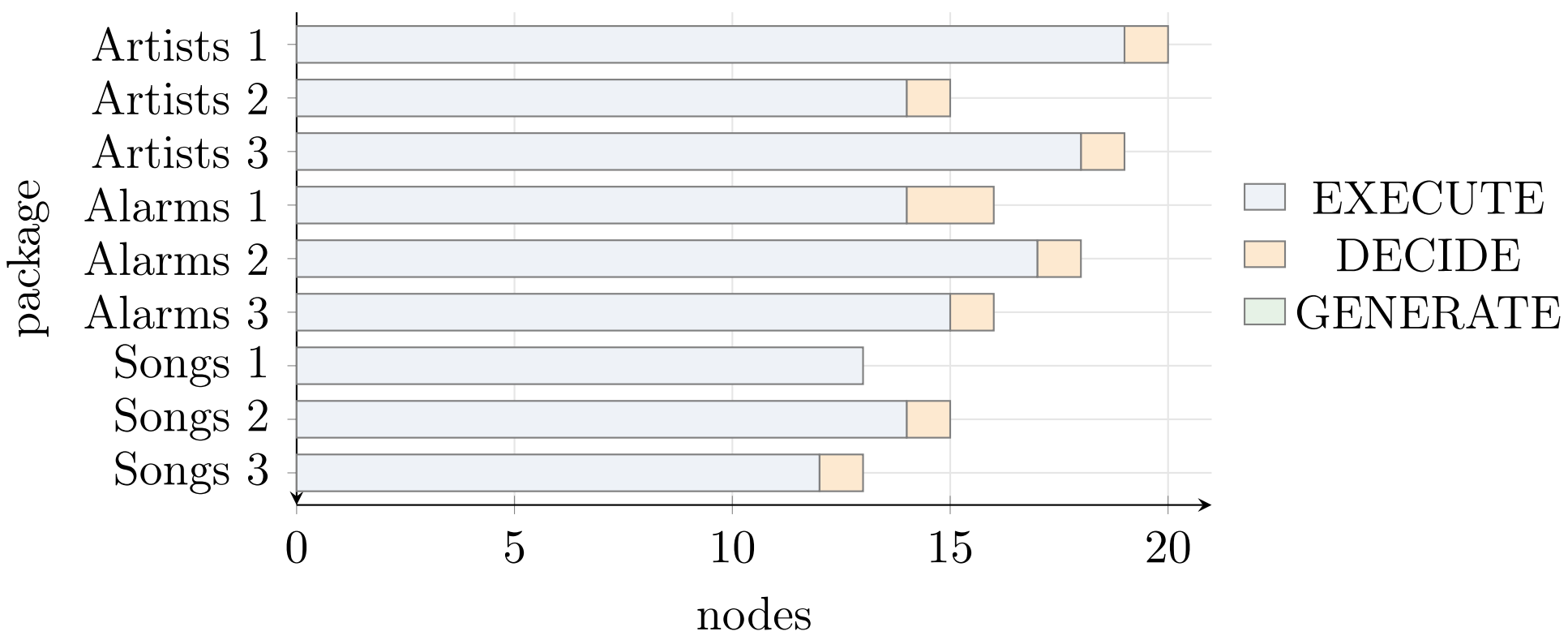

What the recompiles do show is that the same traces admit different graphs. Figure~\ref{fig:composition}
splits every package into EXECUTE, DECIDE and GENERATE nodes: the nine range from 13 to 20 nodes, and
the same family compiled three times gives 20, 15 and 19 nodes for Artists and 13, 15 and 13 for
Songs. The composition moves with the size: the two Songs recompiles read the comparison of the
sentence with a decision where the main package parses it in code, and the alarm recompiles trade one
decision for a longer chain of operations. None of these differences changes the outcome: the workflow is a family of
equivalent programs, and the measurement survives the draw.

\subsection{Ablations}
\label{sec:ablations}
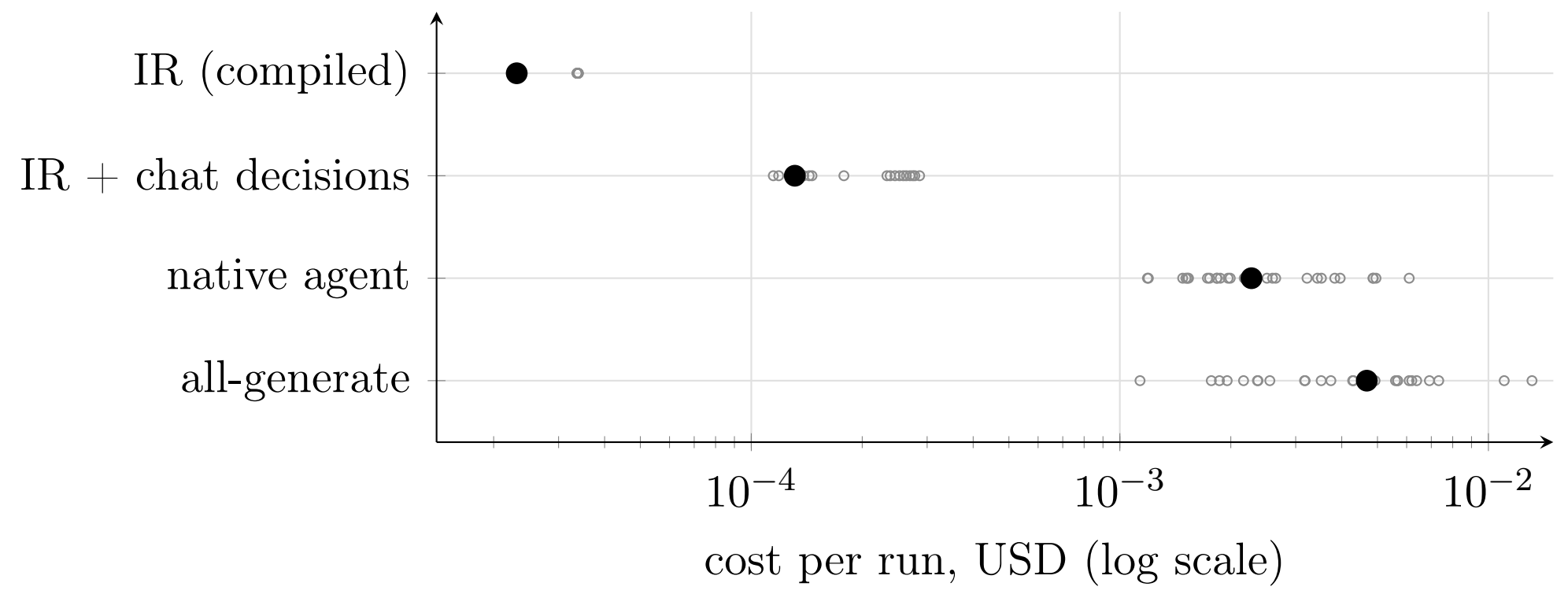
\begin{figure}[tbp]
\centering
\begin{tikzpicture}
\begin{axis}[
  width=0.80\textwidth, height=5.2cm,
  xmode=log, xmin=1.4e-5, xmax=1.5e-2,
  ymin=0.4, ymax=4.6,
  ytick={1,2,3,4},
  yticklabels={all-generate, native agent, IR + chat decisions, IR (compiled)},
  xlabel={cost per run, USD (log scale)},
  tick label style={font=\small},
  label style={font=\small},
  y tick label style={font=\small, align=right},
  grid=major, grid style={black!12},
  axis y line=left, axis x line=bottom,
]
\addplot[only marks, mark=o, mark size=1.1pt, draw=black!45, fill=black!15, forget plot]
  table[x index=1, y index=0] {figures/cost-instances.dat};
\addplot[only marks, mark=*, mark size=2.4pt, draw=black, fill=black]
  table[x index=1, y index=0] {figures/cost-medians.dat};
\end{axis}
\end{tikzpicture}
\caption{Cost of one held-out instance in every arm: a dot is a run, the filled marker the median of thirty.
The two arms that carry context sit three orders of magnitude above the two that do not; a run with
no model call cannot appear on a log axis, so the ten such runs of Songs are in the median only.}
\label{fig:strip}
\end{figure}

The comparison of Table~\ref{tab:main} changes two things at once: it replaces the agent's loop with
an explicit graph, and it replaces model calls with computation. Table~\ref{tab:ladder} puts four
arms on the same 30 held-out instances, at the same prices and under the same step budget, one run
per instance with nothing retried, so that the difference between two rows is attributable to what
they change. Figure~\ref{fig:strip} plots every run of the four arms, so the medians the table
reports can be read against the spread. The native agent is the loop as it runs, with no
artifact; the IR row replaces that loop with the compiled package; the row above it keeps the package
and changes only who answers the decisions, from the typed decision model to a large generative one;
and the all-generate row removes the package altogether and asks whether the traces alone can carry
the procedure.

Read from the bottom, the ladder says that typing a decision is worth little and that removing it is
worth a great deal. Putting the same packages' DECIDE nodes to the large generative model, as
labelled questions instead of to the typed decision model, multiplies the per-run cost by about
seven, from \$0.000019 to \$0.000132, and leaves success unchanged. The questions are
rare, one per run in one family, two in another and none in the third, so the size of the decision
model is not where the saving comes from; even at seven times the price of a typed decision, the
compiled workflow remains twenty times cheaper than the agent, because there is almost nothing left
to ask once the EXECUTE nodes have become code.

Read from the top, the ladder bounds the alternative. The all-generate arm puts every trace of the
family into one prompt, appends the task, and executes the python of a single reply against the live
world; it solves 30/30, so the traces alone carry enough information to write a procedure that works
on worlds the traces never saw, at this resolution. It is not a cheap way to carry it. One call that
rereads the traces costs \$0.008056 per run, three times the agent's \$0.002644 and 425 times the
compiled package's \$0.000019, because the demonstrations are paid for again on every instance, and
the prefix cache that the arm's ordering already exploits answers nine tenths of its input. On
the means of Table~\ref{tab:ladder}, the agent is 139 times the compiled package for one fewer
success. Pricing every input token at
the miss rate leaves the order unchanged: \$0.038451 and \$0.029585 per run for the two arms that
carry context, against \$0.000019 for the package, so the conclusion does not depend on caching. What
caching changes is the size of the gap, by factors of five and eleven, which is why the table reports
both.

Two arms that the design note for this table named are absent. A procedural script, which would need
no inference anywhere, would have to be written per family by a person, and we did not write one. A
structure without lowering, in which the graph is kept and every node including EXECUTE is answered
by a model, is not a valid package under the rules of Section~\ref{sec:ir}, and we did not build it.
Neither would change what the two rows that share a graph already show: on the same nodes, the choice
of decision model moves the cost by seven and the success by nothing, while replacing the agent's
loop with the compiled package moves it by two orders of magnitude.

\begin{table}[tbp]
\centering
\footnotesize
\setlength{\tabcolsep}{4pt}
\begin{tabular}{@{}lcrrrrrc@{}}
\toprule
Arm & Solved & In tok & Cached & \$/run & \$ uncd & s/run & Calls \\
\midrule
All generate & 30/30 & 115{,}076 & 103{,}386 & 0.008056 & 0.038451 & 16.4 & 1.0 \\
Native agent & 29/30 & 90{,}021 & 79{,}258 & 0.002644 & 0.029585 & 46 & 10.1 \\
IR, large-model decisions & 30/30 & 255 & 38 & 0.000132 & 0.000143 & 1.5 & 1.0 \\
IR (compiled) & 30/30 & 452 & 0 & 0.000019 & 0.000019 & 0.6 & 1.0 \\
\bottomrule
\end{tabular}
\caption{The ladder, on the same 30 held-out instances and price table as Table~\ref{tab:main}.
IR, large-model decisions runs the same packages and puts each decision to DeepSeek V4.1 Flash; all
generate makes one call per task with every trace in the prompt. Tokens are per run, averaged over
the 30 instances; the two price columns differ in whether cached input is charged at the cache rate
or the miss rate.}
\label{tab:ladder}
\end{table}

The token columns make the same point in the invariant quantity. The two arms that carry context
reread it on every instance: the agent sends 90,021 input tokens per run, of which the provider
answers 88\% from its prefix cache, and all-generate 115,076, of which 90\%. The compiled package
sends 452, which is the sentence and the candidate list of one or two decisions; the rest of its
work is code. The arm that puts the same decisions to a large model sends 255, so the sevenfold
price between the two bottom rows is the model that answers. What
separates the compiled arm from the two that carry context is the removal of the context itself,
which is why the gap is two orders of magnitude at comparable success.

 \section{Discussion}
\label{sec:discussion}
This section returns to the results and discusses what they imply, in the order in which they were
presented.

\subsection{When compilation pays}
The economics of the experiment are simple enough to be a decision rule. Building a family's workflow costs $C$, the traces it is compiled from plus the compile, and takes 3 to 5 minutes; a workflow run costs $r$ and an agent run costs $a$, so the investment pays after $C/(a-r)$ executions. With the measured values, $C$ is \$0.110, \$0.063 and \$0.126 for the three families, the saving is the difference between the two mean costs of Table~\ref{tab:main}, and the thresholds are 50, 41 and 31 executions. Two consequences follow. The threshold is in the tens of executions per request, not in the thousands, because the agent is two orders of magnitude more expensive per run than the workflow; and the threshold is set by what the artifact costs to build, traces and compile, not by the workflow's own price, so a further tenfold reduction in the workflow's per-run cost would move the threshold by less than a tenth. The rule also says when not to compile: a request performed a handful of times is cheaper to leave to the agent.

Read against the instance supply of this study, the numbers are tighter than the rule suggests. Each family produced between 24 and 27 instances, of which 10 are held out and the rest traced. If a deployment saw each instance once, no workflow would repay its investment within that supply: the song workflow comes closest, at 31 executions against 27 instances, while the alarm workflow would need 41 against 24 and the artist workflow 50 against 25. The rule is the result, and the coincidence is that it lands within the size of the family, which is why how often a request recurs matters more than how cheap the workflow is per run. A deployment of these tasks would have to see its requests more than once for the compilation to pay,
and that is the premise the paper started from.

Once the compile is paid, what a run still costs is one or two decision calls, 452 input tokens on
average, and that is the whole of the residual: Table~\ref{tab:ladder} gives the accounting
per arm.

\subsection{What lowers and what does not}
The category that moves cost is DECIDE, but the compiled packages use it as a typed extractor, not a router. In the compiled alarm workflow both decision nodes route every one of their labels to the same successor, and the chosen label is read by an EXECUTE node that compares it with the world and routes on the result; in the artist workflow the single decision declares a linear successor and its answer is consumed downstream. The saving therefore does not come from turning an if-statement into a model call. It comes from turning free text into a label over a set of candidates that the system enumerates, which is a smaller question than the sentence and can be answered by a decision model at a fraction of the price of an agent turn. A compiler that had routed on model answers would have moved the same decisions into the graph's control flow and bought nothing.

The family that stops at no model calls and the families that stop at one or two also mark the boundary of what the IR lowers. A parameter whose value the sentence determines can become an operation: a year, a comparison operator, an amount. A parameter that names an entity of the world cannot, because the world's entities are not known at compile time, and asking a model to produce one would be a GENERATE that invents identifiers instead of reading them. The IR's answer is a decision whose criteria can come from state, which is what the genre decision in these packages does; the two alarm decisions use fixed criteria, and the direction is a two-label choice that the compiler left to the decision model, which a different compiler could have parsed.

Whether a sentence that has to be read becomes a parser or a generation step is a judgement the
compiler makes, not a property of the step. The cutoff year of the song family is a regular
expression in the main package, a decision over four labels for the comparison in one recompile, and
a GENERATE node in two of the coverage compiles, and every one of those packages passes. The
judgement fails in a predictable direction when traces are scarce: with one trace the parser is
fitted to the phrasing of that trace, which is the failure of the one-trace packages. GENERATE is
the fallback that does not depend on the phrasing, at one model call per run. This is also why the
three main packages contain no GENERATE node and five of the twelve coverage packages do: the
boundary moves with the evidence, and we report what each compile produced instead of a canonical
graph. The outputs of those nodes are integers, so a decision over the years the library contains
would have been the tighter lowering by the rule of Section~\ref{sec:ir}; the compiler took the
looser one, and the packages pass, which is the escape hatch used as intended.

\subsection{Where the saving comes from}
The ablations separate the two things that the comparison changes at once. Typing the decisions is
worth little: putting the same DECIDE nodes to a large generative model multiplies the run by seven
and changes nothing else, because the questions are one or two per run and there is almost nothing
left to ask once the EXECUTE nodes are code. Removing the agent's loop is worth everything: the package
replaces it and is 98 times cheaper per run, and that is where the two orders of magnitude live. The all-generate arm closes the alternative from the other side: the traces alone
carry enough information to write a procedure that works on unseen worlds, but carrying them in a
prompt costs three times the agent and 425 times the package, because the demonstrations are paid for
again at every instance. The saving is in the lowering, not in the size of the model that answers
what is left.

\subsection{Recent compilers}
Two recent systems make the same bet with a different artifact. TraceCompiler mines clusters of
traces into a DAG in which a model survives only in the bindings that the observed values could not
resolve, and refuses an intent whose side effect is under-determined \citep{tracecompiler2026}.
FlashScript writes a neuro-symbolic program from a run, with typed model calls at fixed points, at a
third of the runtime of a skill-guided agent \citep{flashscript2026}. Both localize the model, and both
report what a run costs without reporting what the compilation cost. Our
form differs in what a node may be: a decision declares its candidates before the run, an answer
outside them is a runtime failure, and control flow is owned by the engine rather than by the code
around the call. Section~\ref{sec:ablations} shows why the difference is not cosmetic: on the same graph,
answering a decision with a generative model instead of a typed one multiplies the cost by seven
and changes nothing else, while removing the decisions removes the cost.

 \section{Conclusions}
\label{sec:conclusion}
In this work, we have proposed to compile successful agent traces into typed workflows. The
intermediate representation has three node categories, EXECUTE, DECIDE and GENERATE, defined by the
output space of a step, and that definition decides whether a node runs as code, as a bounded
decision or as a generation, and therefore what a compiled artifact costs and what it can verify. A
compiler agent reads the traces of a family, the IR specification, the deployment document and a
world of only the traced instances, and writes a package that the engine runs without a model
wherever the graph does not need one.

We tested the approach on three families of AppWorld and 30 held-out instances. The compiled
workflows solve 30/30 at a median \$0.000023 and 0.6 seconds per run, against 29/30, \$0.002276 and 46
seconds for the agent; one family compiles to no model call, three traces sufficed for the two that
need decisions under this split, and the compilation is repaid after 31 to 50 executions. The artifact is also
inspectable in a way that a prompt is not: the routes are declared, every decision names its
candidates, and the model calls can be counted before the run.

The evidence has limits. The split withholds the world while the request stays visible, so the
measured generalization is over databases, users and dates; ten instances
per family bound the success comparison, while the cost and latency differences are counted, not
estimated; and one compiler and one benchmark produced the packages, with every trace a success.
Interaction with the user is specified but not measured, and the GENERATE category is exercised only
in the coverage compiles, where it reads a parameter from the instruction. Prices were read on
21 October 2026 and wall clock on one machine, so the dollars follow a published table and the
seconds should be read as an order of magnitude.

The results support the premise with which we started, that a recurring request can be compiled once
and executed cheaply many times, and they leave three lines of work. The first is a selector: given a
new request, retrieve the traced families that resemble it, decide whether one of them covers it, and
compile or reuse the workflow that applies. The second is a dispatcher at run time: send each request
to the compiled workflow when one applies and to an agent when none does, with both sharing the task
record and the completion convention. The third is the compiler itself, where optimiser passes could
lower a graph further once it runs, turn a GENERATE that the new traces now support into a decision
or a parser, and shrink the packages that the current compiler leaves larger than they need to be.
None of the three is measured in this paper.
 
\section*{CRediT authorship contribution statement}
Jorge Garc\'ia-Carrasco: Conceptualization, Methodology, Software, Validation, Investigation,
Data curation, Writing -- original draft, Writing -- review \& editing, Visualization. Javier
Sanchis: Conceptualization, Methodology, Validation, Investigation, Data curation, Writing --
original draft, Writing -- review \& editing, Visualization. Sandra Garc\'ia-Ponsoda:
Conceptualization, Methodology, Validation, Investigation, Data curation, Writing -- original
draft, Writing -- review \& editing, Visualization. Alejandro Mat\'e: Supervision, Project
administration, Conceptualization, Writing -- review \& editing. Juan Trujillo: Supervision,
Project administration, Conceptualization, Writing -- review \& editing.

\section*{Declaration of competing interest}
The authors declare that they have no known competing financial interests or personal
relationships that could have appeared to influence the work reported in this paper.

\section*{Data availability}
The experiment is a set of JSON and JSONL records: the 46 traced sessions, the compiled packages
with their graphs and operations, the held-out runs of every arm, and the usage record of every model
call that the cost figures are recomputed from. They are archived at
\url{doi:10.5281/zenodo.24537625} under CC BY 4.0.

\section*{Code availability}
The IR library, the compiler harness and the benchmark scripts are at
\url{https://github.com/jgcarrasco/agent-trace-compiler-appworld} (tag \code{v1.0.0}), under
Apache-2.0.

\section*{Funding}
This work was supported by the ENIA Chair of Artificial Intelligence at the University of Alicante
[TSI-100927-2023-6]; the
Spanish Ministry of Science and Innovation under the KOSMOS-UA project [PID2024-155363OB-C43];
and Generalitat Valenciana under the BALIDA-AA [CIPROM/2024/13].

\section*{Declaration of generative AI and AI-assisted technologies in the writing process}
The authors acknowledge the use of DeepSeek-V4.1-Flash for improving the readability and checking the
grammar of this manuscript. The authors reviewed and edited the content as needed and take full
responsibility for the content of the publication.

\bibliographystyle{elsarticle-num}
\bibliography{refs}

\end{document}